\documentclass[lettersize,journal]{IEEEtran}
\usepackage{amsmath,amsfonts}
\usepackage{algorithmic}
\usepackage{algorithm}
\usepackage{array}
\usepackage[caption=false,font=normalsize,labelfont=sf,textfont=sf]{subfig}
\usepackage{textcomp}
\usepackage{booktabs}
\usepackage{stfloats}
\usepackage{url}
\usepackage{verbatim}
\usepackage{graphicx}
\usepackage{cite}
\usepackage{multicol}
\usepackage{multirow}
\usepackage[normalem]{ulem}
\usepackage{makecell,diagbox}  
\usepackage{float}
\usepackage{adjustbox}
\usepackage{hyperref}

\usepackage{tikz}
\usepackage{pifont} % 提供对号和叉号
\usetikzlibrary{matrix, positioning}

\newcommand{\Checkmark}{\textcolor{blue}{\ding{51}}} % 定义对号
\newcommand{\Xmark}{\textcolor{red}{\ding{55}}} % 定义叉号

\hypersetup{
    colorlinks=true,
    linkcolor=blue,
    filecolor=magenta,      
    urlcolor=cyan,
    pdftitle={Overleaf Example},
    pdfpagemode=FullScreen,
    }
\usepackage{tikz}
\usetikzlibrary{shapes.geometric, arrows, positioning, trees, shadows}
\usepackage{forest}
\newtheorem{definition}{Definition}[section]

\begin{document}

\title{A Survey on Contribution Evaluation in Vertical Federated Learning}

\author{Yue Cui,
		Chung-ju Huang,
		Yuzhu Zhang,
            Leye Wang,
		Lixin Fan,
		Xiaofang Zhou, and
		Qiang Yang
		
	\IEEEcompsocitemizethanks{\IEEEcompsocthanksitem Yue Cui, Xiaofang Zhou, and Qiang Yang are with The Hong Kong University of Science and Technology, Hong Kong SAR. Qiang~Yang is also with Webank, Shenzhen, China. E-mail: \{ycuias, zxf, qyang\}@cse.ust.hk
    \IEEEcompsocthanksitem Chung-ju Huang and Leye~Wang are with The Key Lab of High Confidence Software Technologies, Ministry of Education, China, and School of Computer Science, Peking University, Beijing, China. E-mail: chongruhuang.pku@gmail.com, leyewang@pku.edu.cn
    \IEEEcompsocthanksitem Yuzhu Zhang is with Tsinghua Shenzhen International Graduate School, Shenzhen, China. E-mail: yz-zhang22@mails.tsinghua.edu.cn
    \IEEEcompsocthanksitem Lixin Fan is with Webank, Shenzhen, China. E-mail: lixin.fan01@gmail.com
	}
}

\markboth{Journal of \LaTeX\ Class Files,~Vol.~14, No.~8, August~2021}%
{Shell \MakeLowercase{\textit{et al.}}: A Sample Article Using IEEEtran.cls for IEEE Journals}


\maketitle

\begin{abstract}
Vertical Federated Learning (VFL) has emerged as a critical approach in machine learning to address privacy concerns associated with centralized data storage and processing. VFL facilitates collaboration among multiple entities with distinct feature sets on the same user population, enabling the joint training of predictive models without direct data sharing. A key aspect of VFL is the fair and accurate evaluation of each entity's contribution to the learning process. This is crucial for maintaining trust among participating entities, ensuring equitable resource sharing, and fostering a sustainable collaboration framework. This paper provides a thorough review of contribution evaluation in VFL. We categorize the vast array of contribution evaluation techniques along the VFL lifecycle, granularity of evaluation, privacy considerations, and core computational methods. We also explore various tasks in VFL that involving contribution evaluation and analyze their required evaluation properties and relation to the VFL lifecycle phases. Finally, we present a vision for the future challenges of contribution evaluation in VFL. By providing a structured analysis of the current landscape and potential advancements, this paper aims to guide researchers and practitioners in the design and implementation of more effective, efficient, and privacy-centric VFL solutions. Relevant literature and open-source resources have been compiled and are being continuously updated at the GitHub repository: \url{https://github.com/cuiyuebing/VFL_CE}.

\end{abstract}


\maketitle
\section{Introduction}
% In the era of burgeoning data-driven models, the impetus for technological advancement is accompanied by growing concerns about data privacy. As organizations increasingly leverage data for machine learning applications, instances of privacy breaches, including gradients leakage and potential AI manipulations, have garnered heightened attention. The intricate landscape of data protection is shaped by diverse regulations worldwide. 
%[Background of privacy protection and data security] 
In the realm of collaborative learning and intelligence, the accurate assessment of individual contributions is paramount. This process, known as contribution evaluation, is fundamental to the success of any cooperative endeavor, ensuring that each participant's efforts are fairly recognized and that the collective work benefits from the strengths of all parties involved \cite{kwon2022weightedshap,jiang2023fair,rozemberczki2022shapley,liu2022gtg,zhang2022intrinsic}. As we progress into an age where data privacy is of the utmost concern, contribution evaluation becomes even more critical in the context of privacy-preserving machine learning techniques.

In recent years, the rapid advancement of data-driven models has advanced technological progress but also raised significant concerns about data privacy. The proliferation of digital interactions and transactions in our modern society has led to an astounding increase in the generation of personal data. Consequently, there has been a surge in the public's concern for privacy, prompting legislators across the globe to enact stringent privacy protection laws and regulations. In Europe, the General Data Protection Regulation (GDPR) \cite{regulation2018general} emphasizes individual rights and data transparency. The USA navigates privacy concerns through the California Consumer Privacy Act (CCPA) \cite{ccpa}, granting residents control over their personal information. China's Data Security Law (DSL) \cite{dsl} underscores the secure and controllable use of data, restricting cross-border data transfer. Against this background, federated learning (FL) \cite{yang2019federated,nguyen2021federated,lim2020federated,li2021survey} emerges as a critical innovation, offering a privacy-centric approach to machine learning by decentralizing the model training process, thereby reducing privacy risks associated with centralized data storage.

%[VFL and its applications in the industry.] 

\begin{figure}[t]
    \centering
    \includegraphics[width=0.75\linewidth]{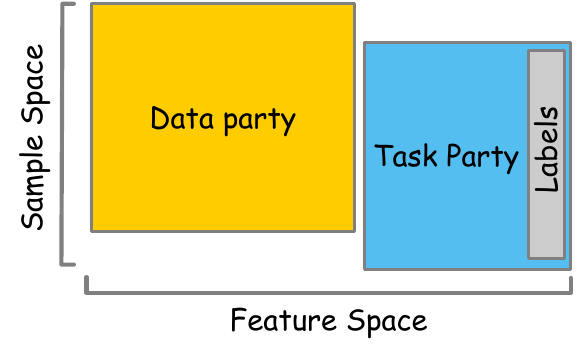}
    \caption{An illustration of the data partitioned in one-on-one VFL setting. }
    \label{fig:space}
\end{figure}

The categorization of FL is primarily based on the partitioning of data across the sample and feature spaces, which results in three distinct forms: Horizontal Federated Learning (HFL), Vertical Federated Learning (VFL), and Federated Transfer Learning (FTL) \cite{yang2019federated,liu2022vertical,saha2021federated}. VFL stands out for its unique ability to enable collaboration among different entities that may have overlapping user samples, yet possess disparate sets of features for the same individuals. A two-party VFL setting is illustrated in Fig \ref{fig:space}. Through VFL, entities can jointly train machine learning models without exposing their raw data, thus maintaining the privacy and confidentiality mandated by law. This approach ensures compliance with privacy regulations while harnessing the collective power of data held across various domains. The role of VFL becomes even more pronounced when considering scenarios where data cannot be pooled due to privacy concerns or logistical challenges, making it a keystone technology in the modern data-centric landscape. The potential applications of VFL are manifold \cite{nevrataki2023survey,wei2022vertical,liu2022vertical}. In the financial world, VFL enables banks and credit agencies to enhance fraud detection systems without sharing sensitive customer information \cite{long2020federated,byrd2020differentially,suzumura2022federated}. In healthcare, hospitals, research institutions, and pharmaceutical companies can collaborate to refine diagnostic tools and develop personalized treatments while preserving patient privacy \cite{antunes2022federated,pfitzner2021federated,chen2020vafl,tang2023ihvfl}. Similarly, in the realm of telecommunications, companies can optimize network operations and customer service by integrating data across different service providers \cite{zhang2020vertical,shome2022federated,niknam2020federated}.

%[Contribution evaluation is important. The success of VFL depends on the quantity and quality of the data owners. ] 
Contribution evaluation plays an important role in this collaborative endeavor \cite{wang2022contribution,wang2019measure,zhang2022intrinsic}. It ensures accountability within the VFL framework. By quantifying and crediting the input of each participant, it establishes a transparent system where the efforts of all contributors are duly recognized. This transparency fosters trust among participants, which is essential for sustaining long-term collaboration. Furthermore, contribution evaluation enables the identification of outliers or potential free-riders within the FL ecosystem. By assessing the quality and consistency of contributions, it helps detect any discrepancies or deviations from expected norms. This proactive approach not only safeguards the integrity of the FL process but also encourages active engagement from all participants. Moreover, contribution evaluation facilitates the allocation of resources and rewards based on merit. Participants who consistently demonstrate high-quality contributions can be incentivized accordingly, motivating them to continue their valuable input. Conversely, those who underperform can receive targeted support or interventions to enhance their involvement, ensuring that the collective effort remains robust and productive. Additionally, contribution evaluation provides valuable insights for optimizing the FL workflow. By analyzing patterns and trends in participant contributions, stakeholders can identify areas for improvement and refine their strategies accordingly. This data-driven approach fosters continuous learning and evolution within the VFL ecosystem, driving innovation and efficiency over time.

%[The current chaos in contribution evaluation definition and taxonomy.]
Though the process of contribution evaluation has been deployed in various VFL applications and tasks \cite{liu2022vertical,sun2023hierarchical}, there is currently no consensus on what constitutes the overarching goal or objective of contribution evaluation in the setting. Furthermore, there is a lack of an established taxonomy or classification system to clearly differentiate and systematically categorize the methods that have been proposed for evaluating contributions in VFL systems. This absence of a uniform, well-defined approach with standardized terminology has led to a chaotic situation where the assessment of contributions is often subjective, inconsistent, and contentious across different VFL implementations. The lack of structured guidance also makes it challenging to objectively compare, select, and improve upon existing evaluation techniques. As such, the establishment of a comprehensive, structured taxonomy for defining and classifying contribution evaluation methodologies in VFL is essential for the advancement of the field.

In this survey, we present an in-depth exploration of existing literature and provide the taxonomy of contribution evaluation in VFL. Inspired by \cite{yang2023federated,kairouz2021advances}, we define contribution evaluation by considering its position in the lifecycle of VFL process: data collection/preprocessing, where data quality and relevance take center stage; model training, which relies heavily on computational contributions and algorithmic efficiency; and model inference, where the practical applicability of the model is tested and the reward of downstream application is allocated. Recognizing the unique contributions in each of these phases is crucial for a comprehensive assessment of the value each participant adds to the VFL collaboration. 

We further investigate contribution evaluation according to its ingredients: the objects and subjects of contribution evaluation, contribution evaluation methods, and tasks related to contribution evaluation. We begin by explicitly summarizing the existing studies according to the objective of contribution evaluation, which can be categorized by granularity into the levels of features and parties. This categorization helps in understanding the target of evaluation, whether it is on individual input features or the collective of a participating party. Since the process of contribution evaluation usually involves information sharing and processing across multiple parties, if looking from the side of the contribution evaluation objective, i.e., the entities that performing contribution evaluation, privacy arises as an important concern. We discuss how existing literature deals with privacy issues from the perspectives of the target data, the contribution calculation agent (the party where the contribution is calculated and compared), and the privacy protection method. It is clear that a delicate balance between accurate contribution assessment and privacy preservation is paramount and the existing strategies aim to strike this balance without compromising either aspect. Then we take the angle of the roadmap of the main techniques, which contains Shapley value based, leave-one-out based, individual based, and interaction based. It should be noted that no single technology is universally superior; rather, the context and requirements of the VFL task should guide the selection of the evaluation methodology. Appropriate adaptation and integration of these techniques can lead to more comprehensive and effective methods for contribution evaluation. Contribution evaluation is also operated differently in different tasks. We delve into details of how it works in feature selection, interpretable VFL, incentive mechanism design, and payment allocation, emphasizing their specific focused properties and methods. These insights illustrate the necessity of a tailored approach to contribution evaluation, one that respects the nuances and objectives of each unique VFL task.

The literature reviewed in the survey was carefully curated from prestigious sources, including articles from IEEE and ACM-affiliated journals and conferences, as well as preprints from arXiv. The most current and authoritative research was included, covering the period from 2016 to 2024. Additionally, the study incorporated comprehensive surveys on FL and VFL published by Springer. We mainly use task-based and method-based keywords for paper searching. The task-based keywords, which include Feature Selection, Fairness, Contribution Evaluation/Valuation, Security, Incentive Mechanism, and Explainability/Interpretability, were chosen for their relevance to contribution evaluation. The method-based keywords, Shapley Value and Leave-One-Out, representing prominent methods used to quantify and analyze the contributions of individual data points or features in a dataset, were also used for searching.

This survey is structured as follows. In Section \ref{sec:pre}, we provide an overview of the fundamental concepts and notations in VFL and delve into the problem formulation of VFL contribution evaluation, highlighting the mathematical ingredients for an evaluation system. Section \ref{sec:tax} presents the proposed taxonomy of contribution evaluation methods within VFL, systematically categorizing these methods across the VFL lifecycle, granularity, privacy awareness, core computational methods, and tasks. Section \ref{sec:cha} addresses the open challenges and future directions in the field of contribution evaluation in VFL. Finally, we conclude the survey in Section \ref{sec:conc}.

\section{Preliminaries}
\label{sec:pre}
This section provides preliminary knowledge of VFL and formulates the contribution evaluation problems in VFL.

\subsection{Federated Learning}
Federated learning (FL) is a distributed machine learning approach that facilitates collaboration among multiple parties while preserving data privacy. The categorization of FL is primarily based on the partitioning of data across the sample and feature spaces, which results in three distinct forms: Horizontal Federated Learning (HFL), Vertical Federated Learning (VFL), and Federated Transfer Learning (FTL) \cite{yang2019federated,liu2022vertical,saha2021federated}.

In the context of HFL, the data across participating entities is characterized by the same feature set but different data samples. This scenario often arises when multiple organizations or devices collect data on the same type of users or subjects but in different environments or under varying conditions \cite{mcmahan2017communication}.

VFL, in contrast, addresses scenarios where the data from different participants aligns with the same samples but features differently across entities \cite{liu2022vertical}. This is particularly relevant when multiple organizations have data on the same individuals but possess distinct information about them, such as financial institutions with access to different aspects of a customer's financial profile.

FTL extends the principles of FL to situations where the data neither fully matches the samples nor completely aligns with the feature spaces \cite{liu2020secure,saha2021federated}. This approach leverages the knowledge gained from related tasks to enhance learning on the local tasks, making it suitable for scenarios where the data is heterogeneous both in terms of samples and features across different participants.

\subsection{Definitions and Notations of VFL}

In the context of VFL, it is customary for each participant to share a common sample space while possessing different feature spaces. There are generally two roles in VFL.

\begin{itemize}
    \item \textit{Task party}, which owns both the features and the labels, and requires intermediate data from other parties to enhance the performance of specific prediction tasks. Consequently, this entity acts as both the initiator and the main executor of VFL processes. The task party is tasked with maintaining and training a top model. This training leverages the local model outputs from participants and its labels. Subsequently, gradients derived from this collective training process are distributed back to the participants. This feedback mechanism enables them to update their local models accordingly. This cyclical exchange underscores the collaborative essence of VFL, where the iterative sharing of knowledge and updates serves to refine and elevate the collective model's predictive accuracy.
    \item \textit{Data party}, which possesses extensive features and trains a local model on private data and sends the output to the task party. The local model is iteratively updated based on the global gradient provided by the task party, thereby ensuring that the learning process is synergistic and reflective of both local and global insights. During the inference phase, the data party still needs to provide collaborative inference services to the task party online.
\end{itemize}

For illustrative purposes, let us consider a simplified scenario with one task party and one data party. An example of the partition of feature space and sample space of the two parties is shown in Fig \ref{fig:space}. Specifically, the task party $t$ maintains a dataset $D_{t}=\{X_{t} \in \mathbb{R}^{|N_{t}|\times F_t}, y\in \mathbb{R}^{|N_{t}|\times 1} \}$, comprising feature space $F_t$ and labels $y$, whereas the data party $d$ holds $D_{d}=\{X_{d} \in \mathbb{R}^{|N_{d}|\times F_d} \}$, characterized by a different feature space $F_d$. These parties may have partially overlapping samples, denoted as $N_o$, where $N_o \subseteq N_t$ and $N_o \subseteq N_d$. The architecture of VFL is ingeniously crafted to enable the collaborative training of a machine learning model without necessitating the direct exchange of data between the parties.

We now consider a general scenario involving $K$ parties, where the entity at position $k_{th}$ serves as the task party. Each party, $D_i \in D_K$, possesses a dataset $X_i$ characterized by the unified sample identifiers $N_o$ but different feature spaces $F_i$, and maintains a local model, $\theta_i$. Additionally, the task party holds a top model $\phi_{t}$ and a set of labels $Y_{t}$, necessitating loss computation based on the outputs from the local models of all parties involved. The primary objective of VFL can be defined as follows.

\begin{definition}
The objective of VFL is to facilitate the collaborative training of machine learning models, denoted as $\{\theta_i\}_{i=1}^K$, to enhance predictive performance for a specific prediction task w.r.t. $Y_{t}$ without the exchange of raw data $X_i$ among the parties by following certain privacy protection constraints, e.g., differential privacy \cite{dwork2006differential}, homomorphic encryption \cite{acar2018survey}, and secret sharing \cite{beimel2011secret}. Mathematically,
\begin{equation}
\begin{aligned}
    \underset{\{\theta_i\}^K_i}{\arg\min}&=\mathbb{L}(\mathbb{F}_{t}(\phi_{t};f_1(\theta_1;X_1),\\
    &f_2(\theta_2;X_2),\cdots,f_{t}(\theta_{t};X_{t})), Y_{t}),\\
    & s.t. \quad \text{privacy protection constraints}
\end{aligned}
\end{equation}

In VFL, it is anticipated that the performance of the federated model surpasses that of models trained locally on isolated datasets, approaching the efficacy of a global model trained on centrally aggregated data. This expectation stems from the collaborative nature of VFL, which amalgamates diverse datasets across various entities, thereby enriching the training process without compromising data privacy. Unlike HFL, which primarily focuses on partitioned datasets with similar feature spaces, VFL is distinguished by its compatibility with a wide array of model architectures. This includes but is not limited to decision trees \cite{WuCXCO20,LiHL0PH0Q22}, linear regression \cite{LiuZKLCHY22}, singular vector decomposition \cite{ChaiWZYC0022}, and neural network \cite{HuangW023, fu2022label}. Such versatility enables VFL to be applied across a broad spectrum of applications, leveraging the unique strengths of different modeling approaches to optimize performance and achieve outcomes that closely mirror those of centralized data training methodologies.

\end{definition}

\section{A taxonomy of Contribution Evaluation in VFL}
\label{sec:tax}
\begin{figure*}[t]
    \centering
    \includegraphics[width=1\linewidth]{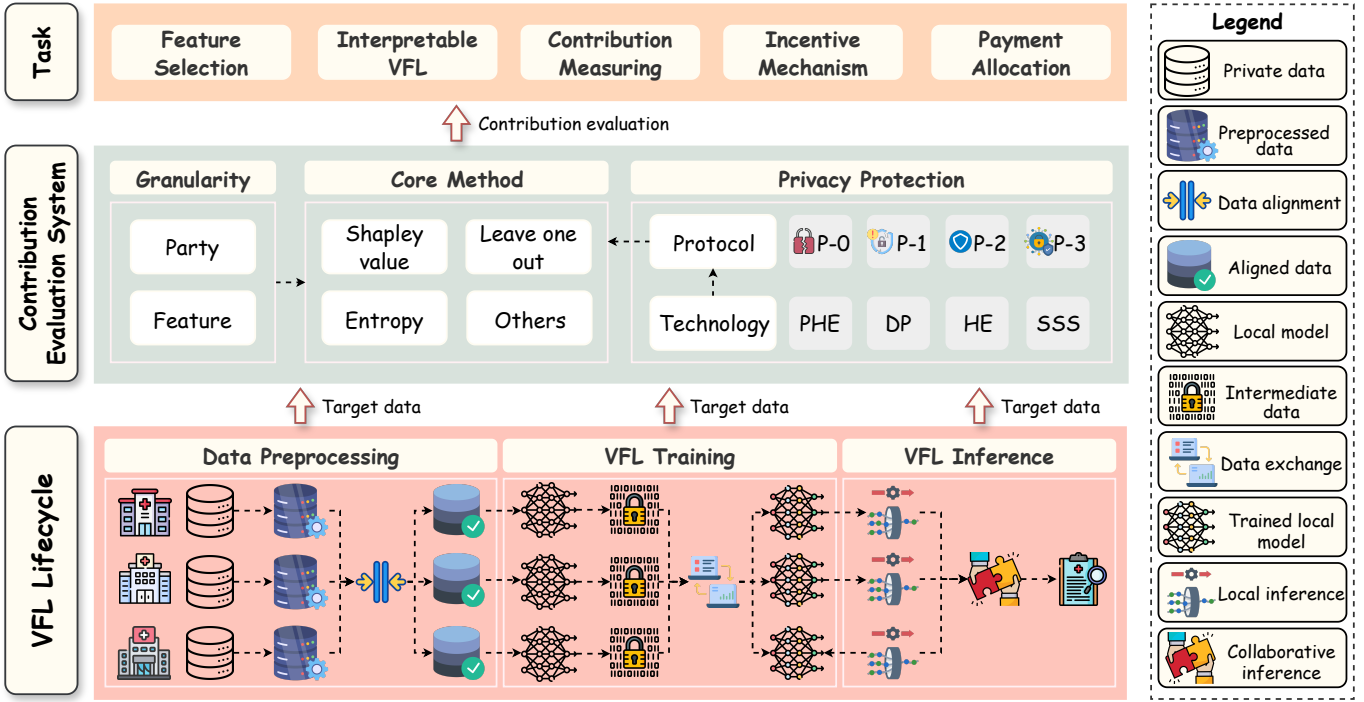}
    \caption{Overview of the structure and function of the contribution evaluation system.}
    \label{fig:overview}
\end{figure*}

\subsection{Problem Formulation}
The mathematical formulation of the contribution evaluation problem in VFL typically involves quantifying the importance or relevance of each participant's contribution to the success of a VFL machine learning model. Suppose there are K parties collaboratively in the training of a model $\phi_{t}$, which is then used to perform a certain downstream task whose quality can be measured by a utility function $\mathcal{U}$. The contribution evaluation problem can be formulated as calculating a contribution score $c_i$ for each participant (or element, e.g., a feature) $i$ that reflects its contribution to the task. Given a contribution evaluation function $f_c$, the computation of $c_i$ can be formulated as:
\begin{equation}
 c_i=f_c(\mathcal{U},\phi_{t},I_t,\{I_i\},E),
\end{equation}
where $I_t$ denotes the information of the task party, $\{I_i\}$, denotes information of data parties, and $E$ denotes potential other external information. Usually, $\mathcal{U}$ is predefined, $\phi_{t}$ is derived from $I_t$ and $\{I_i\}$, and $E$ can be obtained before, during, or after the VFL modeling training. Note that $\mathcal{U}$,$f_c$ and the input of $f_c$ should be adaptive to specific application scenarios. For example, when $c_i$ is expected to be calculated as Shapley value, $f_c$ could be the Shapley value function and $\mathcal{U}$ should be the performance measurement of downstream task and the input of $f_c$ include model performance of possible different permutations in which a coalition with the $i$-th party is formed.

\begin{figure*}
    \centering
    \includegraphics[width=1\linewidth]{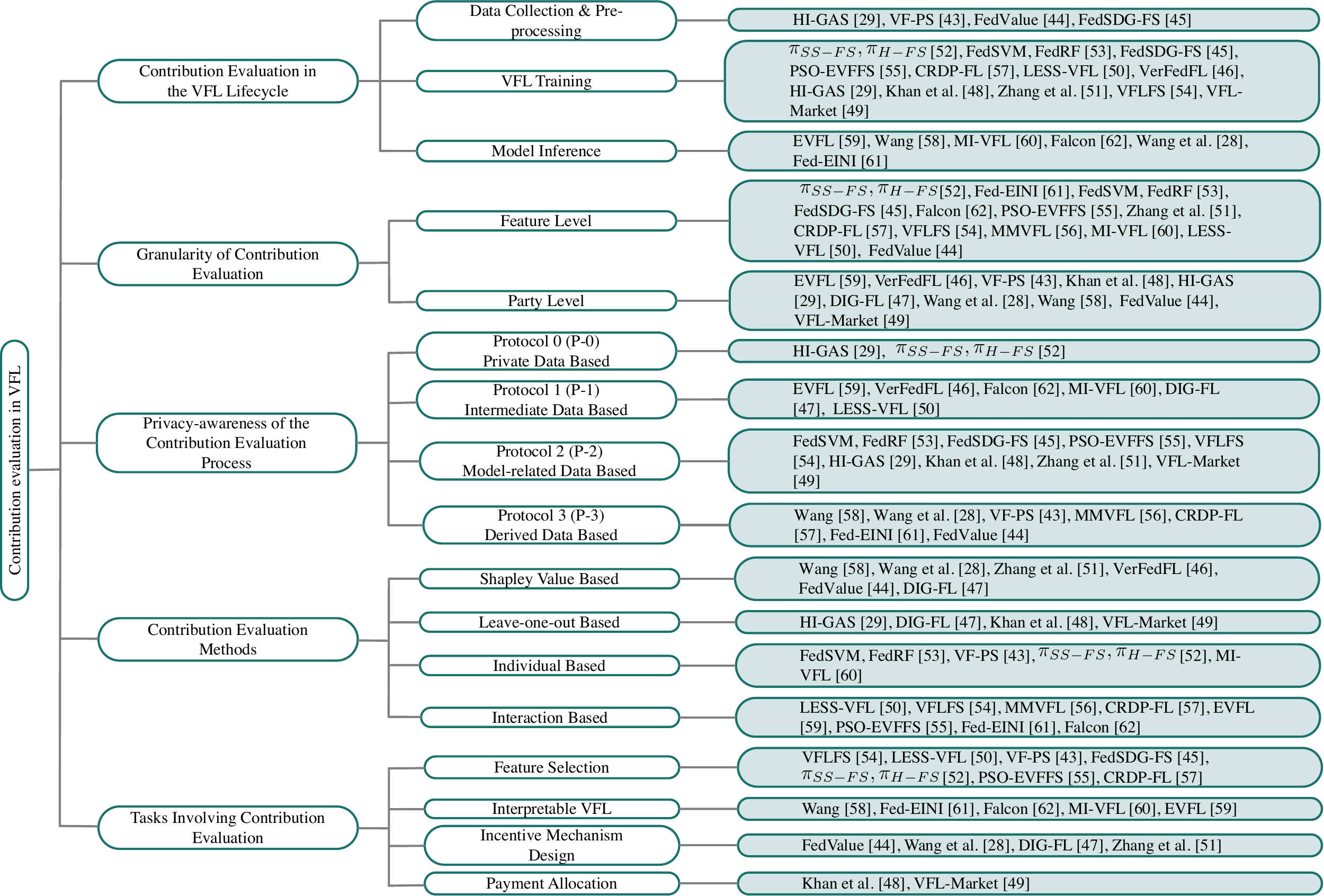}
    \caption{Overview of the proposed taxonomy and corresponding papers.}
    \label{fig:overview_tree}
\end{figure*}

In our survey, we systematically classify the contribution evaluation methods within VFL across five main axes: the VFL lifecycle, the granularity of evaluation objects, the privacy awareness of evaluation subject, the core computational methods, and tasks involving contribution evaluation. Table~\ref{tab:overview} provides a detailed overview, showing different perspectives on the existing research. The lifecycle dimension shows evaluation of contributions plays a key role throughout the learning process, identifying areas where targeted improvements could be made at different stages. The granularity axis explores the different levels at which evaluations are done. This includes evaluations at both large and small scales. We also emphasize the importance of privacy considerations in conducting these evaluations, especially in VFL where sensitive data from multiple parties is involved. Our classification outlines how methods balance accuracy with privacy concerns, and how our specific procedures support different levels of privacy. Then, the core methods axis focuses on the specific algorithms and technical approaches used for evaluations, e.g., Shapley values and informational entropy. This perspective allows for a comparison of how effective these methods are and how they relate to real-world use, helping to choose the most suitable evaluation technologies. Lastly, we discuss the roles contribution evaluation plays in different tasks, emphasising their unique focuses when performing contribution evaluation. Collectively, these aspects form a system for evaluating contributions in VFL that can adapt to different stages of the process and meet the needs of various applications downstream. Fig. \ref{fig:overview} summarizes the structure and functions of the contribution evaluation system. Fig. \ref{fig:overview_tree} shows the proposed taxonomy and corresponding papers.

% \begin{forest}
%   for tree={
%     grow'=0,
%     parent anchor=east, child anchor=west,
%     rectangle,
%     draw,
%     align=center,
%     fill=blue!20,
%     edge path={
%       \noexpand\path [draw, \forestoption{edge}] (!u.parent anchor) -- +(5pt,0) |- (.child anchor)\forestoption{edge label};
%     },
%   },
%   [Contribution Evaluation in VFL, rotate=90
%     [The Lifecycle of VFL
%       [Data Collection \& Preprocessing
%         [HI-GAS \cite{benardSHAFFFastConsistent2021}, edge label={node[midway, left] {}}]
%         [VFL-PS [36], edge label={node[midway, right] {}}]
%       ]
%       [VFL Training, ...]
%       [Model Inference, ...]
%     ]
%     [Granularity of Evaluation, ...]
%     [The Privacy Issue of Contribution Evaluation, ...]
%     [Contribution Evaluation Methods, ...]
%     [Tasks Related to Contribution Evaluation, ...]
%   ]
% \end{forest}

\begin{table*}[t]
\centering
\caption{Summary of existing works on VFL contribution evaluation. (Shapley: Shapley Value Based, LOO: Leave-one-out Based, Individual: Individual Based, Interaction: Interaction Based)}
\fontsize{8.5}{12}\selectfont
\label{tab:overview}
\begin{tabular}{c||c|cc|c|c}
\Xhline{0.8pt}
\multicolumn{1}{c||}{\multirow{2}{*}{Paper}}& \multicolumn{1}{c|}{\multirow{2}{*}{VFL Phase}} & \multicolumn{2}{c|}{Granularity} & \multicolumn{1}{c|}{\multirow{2}{*}{Method}} & \multicolumn{1}{c}{\multirow{2}{*}{Contribution Measurement}} \\
\multicolumn{1}{l||}{} & \multicolumn{1}{c|}{} & \multicolumn{1}{c}{Party} & \multicolumn{1}{c|}{Feature} & \multicolumn{1}{c|}{} & \multicolumn{1}{c}{} \\ \hline
VF-PS \cite{jiang2022vf}         & Data Preprocessing                 & \Checkmark &            & Individual    & Mutual Information   \\ \hline
FedValue \cite{han2021data}         & Data Preprocessing                 & \Checkmark & \Checkmark & Shapley    & Mutual Information \\ \hline
HI-GAS \cite{sun2023hierarchical} & Data Preprocessing, Model Training & \Checkmark &            & LOO    & Model Performance   \\ \hline
FedSDG-FS \cite{li2023fedsdg}        & Data Preprocessing, Model Training &            & \Checkmark & Individual    & Gini Score   \\ \hline
VerFedSV \cite{fan2022fair}         & Model Training                     & \Checkmark &            & Shapley    & Model Performance   \\ \hline
DIG-FL \cite{wang2022efficient}   & Model Training                     & \Checkmark &            & Shapley, LOO & Model Performance  \\ \hline
Khan et al. \cite{khan2023incentive}   & Model Training                     & \Checkmark &            & LOO    & Model Performance   \\ \hline

VFL-Market \cite{cui2024bargaining}   & Model Training                    & \Checkmark &            & LOO    & Model Performance   \\ \hline

LESS-VFL \cite{castiglia2023less}   & Model Training                     &            & \Checkmark & Interaction    & Learnable Feature Weights  \\ \hline
Zhang et al. \cite{zhang2022data}       & Model Training                     &            & \Checkmark & Shapley    & Model Performance  \\ \hline

 $\pi_{SS\_FS}, \pi_{H-FS}$ \cite{zhang2022secure}     & Model Training                     &            & \Checkmark & Indivudual    & Gini Score   \\ \hline
FedSVM, FedRF \cite{ge2022failure}       & Model Training                     &            & \Checkmark & Individual    & Gini Score   \\ \hline
VFLFS \cite{feng2022vertical}    & Model Training                 &            & \Checkmark & Interaction    & Learnable Feature Weights   \\ \hline
PSO‐EVFFS \cite{zhang2022embedded}   & Model Training                     &            & \Checkmark & Interaction    & Node Split Time   \\ \hline
MMVFL \cite{feng2020multi}       & Model Training                     &            & \Checkmark & Interaction    & Learnable Feature Weights   \\ \hline
CRDP-FL \cite{zhao2022vertically}  & Model Training                     &            & \Checkmark & Interaction    & Correlation Score    \\ \hline
Wang \cite{wang2019interpret}   & Model Inference                    & \Checkmark &            & Shapley    & Model Performance   \\ \hline
Wang et al. \cite{wang2019measure}     & Model Inference                    & \Checkmark &            & Shapley    & Model Performance   \\ \hline
EVFL \cite{chen2022evfl}        & Model Inference                    & \Checkmark &            & Individual    & Counterfactual Difference  \\ \hline
MI-VFL \cite{xing2023distributed} & Model Inference                    &            & \Checkmark & Individual    & Learnable Feature Weights   \\ \hline
Fed-EINI \cite{chen2021fed}         & Model Inference                 &            & \Checkmark & Interaction    & Node Split Time   \\ \hline
Falcon \cite{wu2023falcon}        & Model Inference                    &            & \Checkmark & Interaction    & Correlation Score    \\ \hline
\end{tabular}
\end{table*}

\subsection{Contribution Evaluation in the VFL Lifecycle}
The lifecycle of VFL can be effectively divided into three distinct stages: data preprocessing, model training, and model inference. Each phase in the life cycle has distinct characteristics and requirements. Contribution evaluation methods suitable for one phase might not be as effective or relevant for another. By classifying these methods according to the VFL life cycle phases, it becomes possible to adopt the most appropriate and effective techniques for each phase. Furthermore, VFL involves a complex interplay of data, algorithms, and model outputs from multiple sources. Classifying contribution evaluation methods according to the VFL life cycle allows for a comprehensive assessment of contributions at each phase, providing a more complete picture of how each entity's input affects the overall process and outcomes.

\subsubsection{Data Preprocessing} Participants in VFL initially establish the objectives of their collaborative task and work towards a consensus on data, models, and other pertinent elements. This phase primarily comprises two key tasks. The first task involves entity alignment, a process facilitated by encryption-based user ID alignment methodologies, as outlined in prior research \cite{yang2019federated}. This technique confirms the shared users among participating entities, enabling collaboration while ensuring privacy. Notably, during entity alignment, the involved parties remain unaware of the non-overlapping users across their datasets. The second task pertains to data preprocessing. Aligned with the requirements of the VFL task, each participant modifies the aligned training and inference datasets to render them conducive for machine learning model development. Typically, individual parties process and locally store their data to comply with security and privacy constraints.

Commonly used data preprocessing methods at this phase usually involve dimensionality reduction and feature selection. By evaluating the contribution of each feature, VFL can select which features are most relevant to the learning task. This is particularly important in VFL scenarios as parties may not have a complete view of the entire dataset. Each party typically holds a subset of features of the same sample IDs, and not all features might be equally informative. Contribution evaluation provides an efficient way for all parties to understand which parts of the data (i.e., parties or features) are most significant for the model before training, facilitating a better collaborative understanding and trust in the model's decisions. And it helps in this process by highlighting which features to focus on, leading to more efficient computation and potentially reducing the complexity of the learning model.

There have been some studies focusing on how to evaluate contributions during the data preprocessing phase.
A multi-dimensional hierarchical contribution evaluation method was proposed in \cite{sun2023hierarchical} that takes the quality of raw data into account during the data preprocessing phase. It achieves this by integrating a data correlation score with a data quantity score, culminating in a comprehensive assessment of data quality. This evaluation then informs the determination of each participant's contribution. 
The study \cite{jiang2022vf} introduced a vertical joint mutual information estimator capable of estimating the mutual information between participants' features and labels. Then it adopts a group testing approach to identify the most informative subset of participants. This selection process significantly enhances the contribution of the chosen participants to the efficacy of subsequent model training.
Han et al. \cite{han2021data} propose a model-free and privacy-preserving data valuation method for VFL, named FedValue. This method is designed to evaluate the contribution of data parties to predictive analytics tasks during the data preprocessing stage without the need to execute machine learning models. It utilizes an information-theoretic approach based on the Master-conditioned Shapley Value, which accounts for the substitution effect among data from different parties. This approach allows for a fair and efficient assessment of each data party's information contribution to VFL tasks.
Li et al. \cite{li2023fedsdg} propose the federated stochastic dual-gate-based feature selection (FedSDG-FS). The first step of this approach uses Gini impurity in conjunction with partially homomorphic encryption (PHE) to securely and efficiently initialize the importance of features in the data preprocessing phase. This allows for an efficient and secure selection of significant features, contributing to enhanced performance and faster convergence of the global VFL model.

\subsubsection{Model Training} After data preprocessing, participants can start training VFL models locally using aligned samples. This phase primarily comprises two key tasks. The first task involves the training round, which adapts to diverse model requirements inherent in VFL, encompassing linear regression models, tree models, neural networks, etc. This stands in contrast to HFL, where the server predominantly aggregates neural network models from all participating parties. Therefore, the training round is model-specific, and its training details and passed intermediate parameters are determined by the specific model. The prevalent training protocol for federated neural network models involves utilizing gradient descent, necessitating the transmission of local model outputs and corresponding gradients across all participating parties, instead of sharing local data. The second task is training verification. At the conclusion of each training round, all participating parties conduct inference verification on the training model to ascertain whether the outcomes align with the defined objectives. This process includes the selection of participants or features based on the contribution made by the data party during training, aiming to eliminate inactive participants and irrelevant features. Additionally, all parties engage in mutual verification to ensure the authenticity and reliability of the training process. Based on these verification results, decisions are made on whether to continue participation in the VFL task. These two tasks are usually executed alternately multiple times until the training results meet the task requirements.

Contribution evaluation during the training phase generally serves two purposes. The first is to optimize model performance. By evaluating the contribution of each feature or party, VFL models can be optimized for better performance. This evaluation helps in identifying which features or participants are most predictive and should be given more importance during training, leading to more accurate and efficient models.
% Training optimization
Zhang et al. \cite{zhang2022secure} introduce an efficient and privacy-preserving feature selection scheme, particularly in eHealth systems. It designs a general framework based on Gini-impurity to select features that contribute greatly to model training. These protocols allow for private and efficient feature selection in VFL, demonstrating significant accuracy gains in medical datasets while maintaining communication and computational efficiency.
Ge et al. \cite{ge2022failure} propose a presents an empirical study on failure prediction in production lines using federated learning. It improves federated random forest for VFL by introducing feature selection and pruning steps, ensuring that only the most contributive features are selected for the final model. The findings indicate that improved algorithms can achieve comparable results to centralized methods in manufacturing failure prediction.
The second step of FedSGD-FS \cite{li2023fedsdg} mention above is about secure feature selection during training, involving forward and backward propagation on the client and server. The process includes computing and encrypting embedding vectors by clients, which are then processed by the server to evaluate feature contributions for feature selection and to train the model, all while preserving data privacy.
Zhang et al. \cite{zhang2022embedded} propose a method for feature contribution evaluation during the training phase using a particle swarm optimization-based approach. The approach effectively removes irrelevant features, focusing on those contributing most significantly to model accuracy, while maintaining data privacy.
Feng et al. \cite{feng2020multi} propose the multi-participant multi-class VFL (MMVFL) framework, designed for multi-class classification problems in VFL with multiple participants. A feature importance evaluation scheme is also introduced, which quantifies the contribution of each participant's different features to the VFL model and determines which features are prioritized during the training period.
Zhao et al. \cite{zhao2022vertically} propose a method for VFL with correlated differential privacy (CRDP-FL). It introduces a privacy-preserved VFL training approach based on differential privacy between organizations and incorporates feature selection to improve algorithm efficiency and model performance.
Castiglia et al. \cite{castiglia2023less} introduce an embedding-based feature selection method in a communication-efficient manner. The server identifies significant embedding components from each party's pre-trained models. Then each party removes non-significant features locally, based on significant embedding components identified by the server.
The method proposed in \cite{feng2022vertical} simultaneously enforces row sparsity constraints on each dimension of the input data of each party to measure feature importance. It uses the first layer parameters of the model as a regularization function to make feature selection more detailed.

% Contribution evaluation
The second is to measure the real contribution of all parties in the model training phase. Participants not only need to provide high-quality data when participating in VFL training but also consume a large amount of computing resources. Therefore, it is crucial to accurately assess the contributions of all parties during the training phase and establish an effective incentive mechanism. Parties are more likely to engage in the federated learning process when they know that their contributions are recognized and valued. This evaluation can also help in mitigating the risk of free-riding, where some parties might benefit from the model without contributing significantly.
The method proposed in \cite{fan2022fair} for evaluating contributions during the training phase is the vertical federated Shapley value (VerFedSV). This method computes contributions at multiple time points during training, using client embeddings at different timestamps. It addresses the challenges of model dependence in VFL and is designed for both synchronous and asynchronous VFL environments. It can effectively evaluate the contributions of all parties in the training phase without requiring model retraining.
The method proposed in \cite{wang2022efficient} for contribution evaluation during the training phase is named DIG-FL. This approach efficiently estimates the Shapley value of each participant without requiring model retraining. It's applicable to both VFL and HFL. This method allows dynamic adjustment of participants' weights based on their per-epoch contributions, thereby improving the accuracy and convergence speed of the model training.
Sun et al. \cite{sun2023hierarchical} develop a multi-dimensional contribution-aware reward distribution mechanism, evaluating contributions based on data quality and model contributions during data preprocessing and model training respectively. By training local models and assessing prediction accuracy using the symmetric mean absolute percentage error (SMAPE), this approach promotes fairness and encourages participation in VFL training.
The method proposed by Khan et al. \cite{khan2023incentive} for model contribution evaluation during the training phase is based on the concept of the bankruptcy problem. This method calculates the contribution of each party by measuring the improvement in model performance and then distributes the "profit" (or incentive) based on these contributions. This method ensures a fair and efficient distribution of rewards among participants, encouraging their involvement in the federated learning process.
The method proposed by Zhang et al. \cite{zhang2022data} for evaluating participant contributions during the training phase is centered around a Stackelberg game model. It involves calculating the Shapley value for each participant's data, signifying their contribution. This value is then integrated into a data pricing model, where the hosts act as leaders and the guest as the follower. The model accounts for contribution scores and pricing strategies, ultimately determining the optimal data usage and pricing policy for both guests and hosts. This approach balances fair compensation with effective resource utilization in the training procedure.

\subsubsection{Model Inference} During model inference, a trained model is deployed to make predictions on new data. Unlike the verification inference during model training, inference at this stage focuses on new user groups, exhibiting distinct data distributions. In this phase, all participants employ the trained model for collaborative inference on new data, using diverse metrics to holistically assess the model's performance. Furthermore, model inference serves as a cornerstone for generating value in VFL tasks. Therefore, the provision of meaningful explanations regarding inference outcomes becomes crucial, fostering fairer and more transparent profit distribution mechanisms. Contribution evaluation plays a key role in this phase as it enhances the interpretability of the deployed model's predictions. By understanding which features, and consequently, which parties contribute most significantly to a given prediction, users and stakeholders can gain insights into how the model is making its decisions.

There have been some studies discussing how to use contribution evaluation in the model inference phase.
Chen et al. \cite{chen2022evfl} propose an explainable vertical federated learning (EVFL) framework, focusing on data-oriented AI systems. It includes credibility assessment strategy, federated counterfactual explanation, and an importance rate metric for feature importance evaluation. It aims to interpret VFL models while ensuring data privacy, improving data quality, and aiding regulatory understanding.
Wang \cite{wang2019interpret} introduces to balance the model interpretability and privacy using Shapley value. It focuses on revealing detailed feature importance for host features and a unified importance value for federated guest features. The method allows for robust and informative interpretation of FL models, ensuring data privacy while providing insight into the contribution of individual features.
Xing et al. \cite{xing2023distributed} propose a method for model interpretation in VFL named MI-VFL, which addresses feature discrepancy issues. The method aims to accurately identify important features and suppress overlapped ones, thereby improving model performance and interpretability in VFL with misaligned feature spaces.
The method proposed in \cite{wu2023falcon} for feature contribution evaluation in the model inference phase involves a complex process, integrating state-of-the-art interpretable methods in a decentralized setting such as local interpretable model-agnostic explanations (LIME). Wang
\cite{wang2019measure} proposes a feature contribution evaluation method based on Shapley values in model inference phase. The Shapley values are utilized to calculate the importance of grouped features. This measure of feature importance is then used to determine the contribution of each participating party in the FML setup.
The proposed method in \cite{chen2021fed} for evaluating the contribution of input features in VFL is a two-stage framework named Fed-EINI. This approach allows the guest party to understand the meanings of features held by the Host parties without compromising privacy, thus addressing the challenge of interpretability.
\subsection{Granularity of Contribution Evaluation}
Contribution evaluation in VFL can further be discussed according to the granularity of evaluation objectives: individual features or a party as a collection. Therefore, existing research can be categorized into two levels: feature-level contribution evaluation and party-level contribution evaluation.

%\cite{fan2022fair,wang2019measure,wang2022efficient,sun2023hierarchical,chen2022evfl,yan2021fedcm}
\subsubsection{Feature-level}
Feature-level granularity in federated learning contribution evaluation refers to the level of detail at which individual features or variables are considered when participants contribute their updates to a global model. 

Feature-level contribution evaluation can be operated globally or locally. Global evaluation refers to the scenario where a server/host is aware of the features that participated in VFL and then evaluates the contribution of each individual feature on the global model. To achieve so, the server \cite{ge2022failure,zhang2022data,li2023fedsdg,zhang2022embedded,zhao2022vertically,chen2021fed,wu2023falcon} can calculate feature importance globally. For example, Zhang et al. \cite{zhang2022data} simply traverse all possible sorting orders of features to calculate the Shapley Value of each specific feature vector for contribution evaluation. On the other hand, the contribution of features also needs to be evaluated locally on certain participants \cite{xing2023distributed,feng2022vertical,feng2020multi}. \cite{feng2022vertical,feng2020multi} aim at selecting the most representative features by each data party for a corresponding learning task on the task party. As the correlation between features can be found across data parties, feature-level contribution evaluation becomes even more challenging in the local view. It requires algorithms and models to extract meaningful insights considering such correlations when determining the significance of an individual feature. To deal with this problem, MI-VFL \cite{xing2023distributed} makes an attempt to adjust local importance to global importance and suppressing overlapped features among clients.

Confidentiality and privacy are the major concerns in feature-level contribution evaluation, as the individual features are private assets of the task party and the labels of the task party are also locally owned. To this end, feature-level contribution evaluation may pose more severe privacy risks. This is because it involves studying individual data points and their distinctive traits, which can lead to potentially sensitive information. Ensuring that no personally identifiable information is unintentionally exposed during the evaluation process can be a significant challenge, especially given the increasing global attention to data privacy standards and regulations. To deal with this challenge, Castiglia et al. \cite{castiglia2023less} propose LESS-VFL. In LESS-VFL, feature-level evaluation is achieved by using embedding components provided by each data party as a proxy. The task party (referred to as the server in \cite{castiglia2023less}) evaluates if the embedding components from the data parties are non-significant and each data party matches the values of the significant embedding components while removing non-significant features from its model. To deal with anonymous features, Fed-EINI \cite{chen2021fed} conceals decision paths and adapts a communication-efficient secure computation method for inference outputs. MMVFL \cite{feng2020multi} adapts multi-view learning to share label information across multiple participants, but does not preserve data privacy in this process.

% 3) Data Heterogeneity: Heterogeneous data in the context of different parties can pose challenges in feature-level evaluation. For instance, the same feature across different parties might have different distributions, scales, or even interpretations. This data heterogeneity influences how individual features affect the global model, which complicates the process of evaluating contributions at the feature-level. Further, when the number of features varies greatly across different parties, it leads to further complexity in evaluation.

%\cite{han2021data,wang2019interpret,zhang2022data,khan2023incentive,jiang2022vf,xing2023distributed,zhang2022secure,li2023fedsdg,castiglia2023less,feng2022vertical,zhang2022embedded,feng2020multi,zhao2022vertically,chen2021fed,wu2023falcon,cherepanova2023performance,tibshirani1996regression,lemhadri2021lassonet,dinh2020consistent,gorishniy2021revisiting}

\subsubsection{Party-level}
Party-level granularity refers to the level at which contributions or updates are made by individual parties in the VFL system. It especially comes into play in the scenario of multi-party VFL, where multiple data parties contributing to the VFL process. 

Some works measure the contribution of each participant directly by taking the local feature/data of a data party as a whole \cite{wang2022efficient,jiang2022vf,khan2023incentive,fan2022fair,sun2023hierarchical,chen2022evfl,cui2024bargaining}. Jiang et al. \cite{jiang2022vf} formally define the vertically federated participant selection (VF-PS) problem, which selects important participants in vertical federated learning. Khan et al. \cite{khan2023incentive} take the local features of a data party as a whole and evaluates the marginal contribution of a task party to the performance improvement of the federated model. \cite{fan2022fair} embeds the local data of a client into an embedding vector and then computes the significance of each client based on the embedding matrix made up of all clients' embedding vectors. 

Wang et al. \cite{wang2019measure} separately valuate each of the features of a VFL task and takes the sum of a set of data parties' features as the contribution value of the task party. However, Han et al. \cite{han2021data} suggest that the generalization from feature-level evaluation to party-level evaluation is non-trivial, neglecting the feature correlations among data parties is likely to lead to inaccurate data valuation for the parties' collaborative modeling. To deal with this problem, Han et al. \cite{han2021data} propose FedValue, which applies conditional mutual information (CMI) to measure the marginal value of a data party d to the task party given a party joining-order, and compute an average marginal value
by enumerating all data party orders to valuate a party.

% Note that in real applicaiton scenario, there are
% often limited numbers of aligned samples, especially as the number of parties grows \cite{liu2022vertical}. Therefore, VFL is mostly performed between a task party and a few or only one data parties and feature-level contr
%\subsubsection{Not specified}
%\cite{catonFairnessMachineLearning2020,blackLeaveoneoutUnfairness2021,ge2022failure,li2023towards,huangEfficiencyboostingClientSelection2020a,benardSHAFFFastConsistent2021,cheungVerticalFederatedPrincipal2022a,liFeatureSelectionData2018,wang2022contribution}

\subsection{Privacy-awareness of the Contribution Evaluation Process}
In the realm of many industries, adherence to stringent data protection legislations, such as the GDPR \cite{regulation2018general} in Europe and the Health Insurance Portability and Accountability Act (HIPAA) \footnote{\url{https://www.hhs.gov/hipaa/index.html}} in the United States, is imperative. In VFL, data from different parties is not shared or transferred; instead, insights are derived from decentralized data sources. The methodology of contribution evaluation typically refrains from utilizing raw data. However, as several existing works 
 suggest: the exchange of data elements, including labels and gradients, may inadvertently precipitate privacy breaches \cite{zou2022defending, jin2021cafe, luo2021feature}. The susceptibility of various methods to such privacy intrusions varies, underscoring the necessity of a systematic classification based on their proficiency in safeguarding sensitive information. Our analysis of the pertinent literature is anchored on three cardinal criteria to ascertain the extent to which contribution evaluation methods may impinge upon data privacy. The foremost criterion is the \textbf{target data}, which stands as the quintessential determinant in assessing the potential for privacy leakage. This term delineates the specific data utilized in executing contribution calculations. The second criterion, \textbf{contribution calculation agent}, pertains to the entity designated with the responsibility of performing these calculations. This agent may operate on a local level, a centralized server, or through an external party. The same computation results in different privacy risks if it occurs locally and to a third party. The final criterion encompasses \textbf{privacy protection method}, which encompasses a spectrum of techniques employed to shield the target data. Our analysis will categorize and assess the extant literature on contribution evaluation within this domain from these three distinct perspectives.

In an effort to enhance our understanding of privacy awareness across various methods and to establish a standard for future methodologies, we have categorized protocols according to the type of target data they utilize. These classifications include private data, intermediate data, model-related data, and derived data. Accordingly, we have introduced four distinct protocols, designated P-0 through P-3, each tailored to the specific characteristics and requirements of these data categories. This differentiation is imperative, given the unique nature of each data type and the corresponding necessity for bespoke handling and protection protocols.

\begin{itemize}
    \item \textbf{Protocol 0 (P-0)}. P-0 focuses on evaluating contributions based on private data, which is the sensitive and proprietary information held by each participant. This protocol ensures that contribution assessments are directly tied to the unique data each participant brings to the VFL process. However, private data involves sensitive information and must not be shared outside of its original context, requiring strict privacy protection measures. This protocol appears to be at odds with the foundational premises of federated learning, which seeks to collaboratively leverage data while minimizing direct data exposure. Given this fundamental discrepancy, the application of the P-0 protocol for contribution evaluation is not recommended, as it may conflict with the overarching objectives of FL frameworks.
    \item \textbf{Protocol 1 (P-1)}. P-1 employs intermediate data for evaluating contributions. Intermediate data, characterized as the output from initial computational stages, occupies a critical juncture in the data processing pipeline. It requires protocols that ensure integrity and confidentiality while allowing for the necessary computations to proceed. The utilization of intermediate data is particularly insightful, as it illuminates the extent of processing effort and the incremental value introduced through a participant’s data transformation or feature extraction endeavors.
    \item \textbf{Protocol 2 (P-2)}. P-2 protocol systematically evaluates contributions through the lens of model-related data, encompassing parameters, model performance metrics, and other quantifiable indicators of a model's effectiveness. P-2 is ideal for environments where the direct impact on the model's accuracy, efficiency, or performance serves as the basis for evaluating contributions.
    \item \textbf{Protocol 3 (P-3)}. P-3 protocol employs a distinctive approach to contribution evaluation by leveraging derived data, which includes transformed feature data, pseudo-label data, and other forms of processed information that have been further refined from their original state. P-3 is suited to scenarios where the derived insights are the main indicators of value. Moreover, P-3 is distinguished by its emphasis on privacy preservation, which minimizes the exposure of sensitive information by relying on data that has been abstracted away from its original form.
\end{itemize}
The implementation of type-specific protocols offers several advantages. Primarily, it allows for tailored security and privacy measures that align with the sensitivity and usage requirements of each data type, enhancing the overall trustworthiness of the VFL system. Moreover, by addressing the unique needs of each data category, these protocols promote efficiency and effectiveness in data processing and model training, leading to improved performance of the VFL system. This differentiation also aids in compliance with legal and regulatory standards, which often vary based on the nature of the data being handled. Our summary of the privacy awareness of existing contribution evaluation mechanisms is in Table \ref{tab:privacy}.

\begin{table*}[t]
\centering
\caption{Summary of existing works on VFL contribution evaluation. (CC: Contribution Computing, PP: Privacy Protection, HE: Homomorphic Encryption, PSI: Private Set Intersection, PHE: Partially Homomorphic Encryption, SSS: Secret Sharing Scheme)}
\fontsize{8.5}{12}\selectfont
\label{tab:privacy}
\begin{tabular}{c||c|c|c|c}
\Xhline{0.8pt}
Paper                           & Protocol & Target Data              & CC Agent           & PP Method     \\ \hline 
HI-GAS \cite{sun2023hierarchical} & P0       & Label                    & Each party         & /             \\ \hline
$\pi_{SS\_FS}, \pi_{H-FS}$ \cite{zhang2022secure}     & P0       & Label                    & Each Party         & HE            \\ \hline

VerFedFL \cite{fan2022fair}         & P1       & Embedding                & Server             & /             \\ \hline
EVFL \cite{chen2022evfl}        & P1       & Embedding                & Server             & /             \\ \hline
MI-VFL \cite{xing2023distributed} & P1       & Embedding                & Server             & /             \\ \hline
LESS-VFL \cite{castiglia2023less}   & P1       & Embedding                & Local              & /             \\ \hline
Falcon \cite{wu2023falcon}        & P1       & Embedding                & Task Party         & PHE, SSS      \\ \hline
DIG-FL \cite{wang2022efficient}   & P1       & Local gradient           & Server             & /             \\ \hline

HI-GAS \cite{sun2023hierarchical} & P2       & Model performance        & Server             & /             \\ \hline
VFL-Market \cite{cui2024bargaining} & P2       & Model performance        & Task Party             & /             \\ \hline
Khan et al. \cite{khan2023incentive}   & P2       & Model performance        & Task Party         & /             \\ \hline
Zhang et al. \cite{zhang2022data}       & P2       & Model performance        & Server             & /             \\ \hline
FedSVM, FedRF \cite{ge2022failure}       & P2       & Gini impurity            & Server             & /             \\ \hline
FedSDG-FS \cite{li2023fedsdg}        & P2       & Gini impurity            & Server             & PHE           \\ \hline
PSO‐EVFFS \cite{zhang2022embedded}   & P2       & Average Split Times      & Task \& Data Party & Normalization \\ \hline
VFLFS \cite{feng2022vertical}    & P2       & First Layer's Parameters & Server             & /             \\ \hline
Wang et al. \cite{wang2019measure}     & P3       & Federated feature        & Task party         & /             \\ \hline
VF-PS \cite{jiang2022vf}         & P3       & Feature distance         & Task Party         & HE            \\ \hline
Wang \cite{wang2019interpret}   & P3       & Feature ID               & Task Party         & /             \\ \hline
FedValue \cite{han2021data}         & P3       & Sample ID                & Server             & PSI           \\ \hline
MMVFL \cite{feng2020multi}       & P3       & Pseudo-label Matrix      & Local              & /             \\ \hline
CRDP-FL \cite{zhao2022vertically}  & P3       & Feature Frequency        & Local              & DP            \\ \hline
Fed-EINI \cite{chen2021fed}         & P3       & Decision Vector          & Task Party         & HE            \\ \hline
\end{tabular}
\end{table*}

\subsubsection{P-0: private data} Private data encompasses both features and labels. Under the setting of FL, it is a fundamental stipulation that private data must be exclusively retained locally, safeguarding it from external access. This ensures that no party, other than the data holder, has the ability to access this private data. Furthermore, the data holder is prohibited from transmitting this sensitive information to any other entities, thereby maintaining the integrity and confidentiality of the private data within the FL environment.

The data preprocessing phase of the two-step federated evaluation system proposed by Sun et al. \cite{sun2023hierarchical} use correlation coefficients to evaluate the data quality of each party. During the calculation process, this method necessitates the transmission of label data in plaintext by the task party to all participants. This approach inherently poses a risk of privacy breaches. The protocol developed by Zhang et al. \cite{zhang2022secure} introduce a secure feature selection method leveraging secret sharing. In this system, the task party is responsible for encrypting the label matrix homomorphically before transmission to the data party. Upon receipt, the data party computes the feature contribution vector, utilizing the encrypted label data and a local binary matrix which is pre-processed by a defined threshold. While this method does not directly expose private data, it introduces significant additional computational overhead.

% The hybrid protocol based on feature selection designed by Zhang et al. \cite{zhang2022secure} requires each party to compute the derived matrix of features locally. Subsequently, the task party is responsible for determining the feature contribution vector. This process entails participants using homomorphically encrypted label data, provided by the task party, to compute the derived matrix. 

\subsubsection{P-1: Intermediate data} Contrasting with the direct utilization of private data for contribution evaluation, the establishment of an evaluation system centered on intermediate data represents a judicious compromise. Within the context of federated learning, such intermediate data typically encompasses elements like model embeddings and gradients.

Embeddings, in this context, are advanced representations of the original features. They encapsulate complex relationships and patterns within the data, transforming them into a format more conducive to model training and analysis. By examining these embeddings, researchers can discern nuanced insights into how different data sets influence the overall model. In \cite{fan2022fair, chen2022evfl, xing2023distributed}, contribution evaluation is conducted by transmitting local embeddings from all participating parties to a centralized server. Distinctively, the method proposed by Castiglia et al. \cite{castiglia2023less} advocate for local feature selection, utilizing a comparison of the squared difference between current and pre-trained embeddings. In the approach outlined by Wu et al. \cite{wu2023falcon}, a peer-to-peer (P2P) communication framework is established between the task and data parties, enabling the direct transfer of local embeddings to the task party for the computation of the weighted Pearson correlation coefficient.

Gradients, on the other hand, are crucial indicators of how a model responds to each feature during the training process. They essentially reflect the sensitivity of the model to changes in different features. By analyzing gradients, it is possible to ascertain the specific impact of each data point or feature on the model's learning trajectory. In the study presented by Wang et al. \cite{wang2022efficient}, the server employs a distinct approach for evaluating the contribution of participants in a federated learning environment. This method utilizes both gradients and the loss function as key metrics, calculating the participants' contributions at each epoch.

In recent studies \cite{he2023backdoor, yang2023practical, chen2024universal}, it has been demonstrated that the direct transmission of intermediate data in VFL can lead to privacy breaches. The two predominant forms of such privacy intrusions in VFL are feature inference attacks and label inference attacks. Feature inference attack entails a scenario where an adversarial entity, often an external participant in the training process, endeavors to reconstruct the private data of other parties by leveraging shared information within the learning protocol. Sun et al. \cite{sun2021defending} mention that the malicious task party will reconstruct the local original data of the data party based on the embedding submitted by the data party. Jin et al. \cite{jin2021cafe} assume that the server has access to the data parties' model parameters and their gradients. Therefore a malicious server can minimize the difference between the actual gradients from the true data and the gradients from the recovered data. And label inference attack in VFL manifests as a privacy violation where a participating entity, typically with malicious intent, seeks to infer sensitive label information from another party's dataset. Fu et al. \cite{fu2022label} highlight the potential for leakage of training sample labels through the gradient sign. And Qiu et al. \cite{qiu2022your} discuss an attack for inferring labels based on intermediate representations. This approach hinges on the exploitation of the relationship between the representations of samples and their corresponding labels. It involves a meticulous analysis of the distances among various sample representations, utilizing this data to deduce possible relationships or connections. These attacks are particularly pertinent in the context of VFL contribution evaluation. When the contribution calculation is conducted locally by a participant, there exists a potential risk that this participant may acquire additional insights about the data belonging to other parties, thereby facilitating the inference of both features and labels by malicious entities. Alternatively, if the contribution evaluation is managed by a third party possessing labeled data, a malicious third party could potentially exploit the intermediate data to uncover the original features of each participant. The methodologies for contribution evaluation in \cite{fan2022fair, chen2022evfl, xing2023distributed, wu2023falcon, wang2022efficient} exhibit potential privacy risks due to their practice of sharing intermediate data in plaintext with third parties. Contrastingly, the approach proposed by \cite{castiglia2023less} primarily relies on the analysis of embeddings at different epochs, which are exclusively owned by the participants. This method inherently bears a lower risk of privacy leakage, as it avoids the external sharing of sensitive data in an unencrypted format.

\subsubsection{P-2: Model-related data} In contrast to traditional methodologies, the use of model-related data in VFL contribution evaluation offers enhanced insights into the impact of each participant's data on overall model performance. It helps in objectively quantifying the contribution of each party. Further, such evaluations can be categorized into two distinct types: black-box evaluation, which focuses on external performance indicators, and white-box evaluation, which delves into the internal mechanics of the model. This bifurcation allows for flexibility in choosing the evaluation method based on the desired balance between transparency and data privacy.

Black-box evaluation involves assessing contributions based on external metrics. This method involves assessing a system's functionality without knowledge of its internal workings. The evaluation is based solely on the input-output relationship. These indicators provide an overview of the model's effectiveness without revealing its internal workings, thus maintaining data privacy and security. In the studies conducted by \cite{sun2023hierarchical, khan2023incentive, zhang2022data}, the methodology for evaluating contributions involves a third party who assesses only the performance of the participating models. Crucially, these third parties do not have access to in-depth model details, nor to the raw or intermediate data utilized in the model's training and operations. This approach ensures a level of privacy and security by limiting the information available to the third parties to just the performance outcomes of the models.

Contrary to black-box, white-box evaluation involves a thorough examination of the internal logic and structure of the system. It offers detailed insights into how different features and components contribute to the model's performance. Ge et al. \cite{ge2022failure} use the local Gini coefficient for determining the optimal feature in a federated learning context. Li et al. \cite{li2023fedsdg} implement local Gini impurity for feature selection, focusing on each feature's purity in classification tasks. Zhang et al. \cite{zhang2022embedded} adopt a different approach, using XGBoost locally to count feature split times, with the task party subsequently determining feature importance based on these averaged split times. Although these methods provide valuable insights into feature relevance, they also pose privacy concerns. Gini impurity and split times can reflect the distribution of original features to a certain extent, and may theoretically be used to infer or restore some features of the original data. Ge et al. \cite{ge2022failure} do not encrypt the Gini impurity, while Li et al. \cite{li2023fedsdg} employ PHE for enhanced security. Zhang et al. \cite{zhang2022embedded} normalize the split times of features held by each participant to prevent potential data distribution analysis by third parties. The feature selection methodology introduced by Feng \cite{feng2022vertical} employs the initial layer parameters of models to ascertain feature contributions, incorporating these evaluations into the loss function via a regularization term to mitigate redundancy. 
%Yan \cite{yan2021fedcm} utilizes an attention mechanism within their framework to assess each participant's contribution, where the server is granted access to the local models in plaintext. However, this transparency poses significant privacy risks, as it potentially enables malicious entities to reconstruct private data from the shared model parameters. To mitigate these vulnerabilities, it is imperative to incorporate advanced security measures such as homomorphic encryption, secure multi-party computation, or trusted execution environments (TEE). 

Black-box contribution evaluation assesses participant/feature contributions without accessing the model's internal mechanics, focusing instead on output or performance metrics. This approach ensures high privacy and security levels, as it does not require revealing the model's architecture or parameters. On the other hand, white-box contribution evaluation delves into the model's internal workings, using detailed information like model parameters, weights, or architecture for a comprehensive analysis. While this method can offer more precise insights into contributions, it poses higher privacy risks due to the exposure of sensitive model details.

% \begin{itemize}
%     % \item \cite{sun2023hierarchical}, model performance, server
%     % \item \cite{khan2023incentive}, model performance, task party
%     % \item \cite{zhang2022data}, model performance, server
%     % \item \cite{ge2022failure}, gini impurity, server
%     % \item \cite{li2023fedsdg}, gini impurity, server
%     % \item \cite{zhang2022embedded}, Average Split Times, task party
%     % \item \cite{feng2022vertical}, First Layer’s Parameters, Local
%     % \item \cite{yan2021fedcm}, model parameters, local
% \end{itemize}

\subsubsection{P-3: Derived data} Leveraging derived data for contribution evaluation in VFL offers a privacy-preserving alternative to sharing sensitive raw data. Derived data, produced through transformation or aggregation of raw data, minimizes privacy risks by transmitting only non-identifiable information. This methodology ensures the privacy of individual data points while still yielding critical insights into each participant's contribution to the collective model, facilitating a fair assessment within the bounds of strict privacy regulations.

In \cite{wang2019measure, wang2019interpret}, the private features are combined into one federated feature and the Shapley value of the federated feature is calculated with the features of all other parties, instead of giving out individual Shapley values for all features in its feature space. In the work presented by Jiang et al. \cite{jiang2022vf}, participants engage in a distance calculation between local and query data, subsequently encrypting these distances via HE before transmission to the aggregation server. The aggregation server consolidates these distances, and the task party utilizes the aggregated results alongside labels to discern the participants' contribution levels through mutual information. MMVFL \cite{feng2020multi} constructs an embedding matrix, where each row is the representation of the corresponding data point. This matrix, serving as a pseudo-label for original features, incorporates regularization terms and orthogonality constraints, thus enhancing the interpretability of feature contributions. Meanwhile, Zhao et al. \cite{zhao2022vertically} propose a distinct feature contribution methodology that computes the ratio of feature frequency to the number of data subsets, identifying significant features based on a predetermined threshold. The interpretability framework established by \cite{chen2021fed} determines the decision vector based on whether the leaf node is in the candidate set. The task party aggregates the homomorphically encrypted decision vectors received from all parties to yield interpretable inference results. Han et al. \cite{han2021data} converts federated Shapley-CMI calculation into calculating the intersection cardinality between parties in a manner that preserves the privacy of the original data. To facilitate these calculations on servers where trust may be an issue, they have devised a dual-server-aided private set intersection (PSI) cardinality computation mechanism, enhancing the security and privacy of the FL process.

\subsection{Contribution Evaluation Methods}

Primary methods for contribution evaluation in VFL are generally categorized into four groups in this subsection. The first group, Shapley Value-based methods, are rooted in cooperative game theory and assess contributions by examining the marginal effect of each participant across all possible coalitions. The second group, Leave-one-out (LOO) methods, streamline the evaluation by measuring the change in performance when omitting individual from the coalition, which provides efficiency but may raise fairness concerns. In contrast to the first two groups, which focus on the impact of excluding an individual from a coalition on model performance, the third group, individual-based methods, evaluates contributions from the perspective of each participant's local data. These methods employ metrics such as mutual information or the Gini coefficient to quantify the value of individual contributions. Lastly, interaction-based methods evaluate contributions by comparing the local data of each participant through metrics such as learnable feature weights and correlation scores, which facilitate an interactive analysis. Table \ref{tab:complexity} provides a summary of the computational efficiency and fairness analysis for the methods. In the subsequent discussion, we will delve into each category of methods in detail.

\subsubsection{Shapley Value Based}
The Shapley Value \cite{shapley1953value} is derived from the field of cooperative game theory, offering a way to fairly distribute the payoff among the players based on their contribution. The Shapley Value for each player is calculated by considering the marginal contributions of every possible combination of players. In this way, it provides a fair evaluation of the contribution of each participating node towards the overall learning model.

Given a universal set of parties/features as $\mathcal{I}$, denote a subset of $\mathcal{I}$ as $S$  and a party/feature as $i$, where $S\subseteq \mathcal{I}\setminus {i} $, an utility function $\mathcal{U}$, the Shapley value of $i$ uder utility function $\mathcal{U}$ is calculated as 
\begin{equation}
\begin{aligned}
v(i)&=\sum_{S\subseteq \mathcal{I}\setminus \{i\} }\frac{|S|!(|\mathcal{I}|-|S|-1)!}{|\mathcal{I}|!}(\mathcal{U}(S\cup \{i\})-\mathcal{U}(S))\\
&=\frac{1}{|\mathcal{I}|}\sum_{S\subseteq \mathcal{I}\setminus \{i\} }\binom{|\mathcal{I}|-1}{|S|}^{-1}(\mathcal{U}(S\cup \{i\})-\mathcal{U}(S)),
\label{eq:sv}
\end{aligned}
\end{equation}
where $|S|$ is the size of subset $S$, $\binom{a}{b}$ is the  binomial coefficient. The Shapley value possesses several desirable properties (a.k.a. fairness):  
\begin{itemize}
\item \textbf{Additivity (a.k.a., Efficiency/Group Rationality)}:
\begin{equation}
\sum_{i \in \mathcal{I}} v(i) = \mathcal{U}(\mathcal{I}).
\label{eq:fairness}
\end{equation}
This property ensures that the total value allocated to all parties is equal to the total value of the grand coalition, promoting a fair distribution of the total worth.

\item \textbf{Symmetry}:

If $\mathcal{U}(S\cup \{i\})=\mathcal{U}(S\cup \{j\})$, $S\subseteq \mathcal{I}\setminus \{i,j\}$, then
\begin{equation}
v(i) = v(j) .
\label{eq:symmetry}
\end{equation}
Symmetry implies that parties/features with the same contribution to any coalition should receive the same Shapley value.

\item \textbf{Linearity}:

Denote the Shapley value under utility function $\mathcal{U}$ as $v_{\mathcal{U}}$, and that under another utility function $\mathcal{W}$ as $v_{\mathcal{W}}$, linearity can be expressed as follows, 
\begin{equation}
v_{\mathcal{U}+\mathcal{W}}(i)=v_{\mathcal{U}}(i)+v_{\mathcal{W}}(i).
\label{eq:linearity}
\end{equation}
This property reflects the linearity of the Shapley value with respect to changes in the utility function.

\item \textbf{Zero Element}:
\begin{equation}
v(\emptyset) = 0.
\label{eq:zero_element}
\end{equation}
The Shapley value for an empty element is zero, indicating that a party/feasture contributes nothing to any subset of the federation receives no value.
\end{itemize}

These properties collectively make the Shapley value a compelling solution concept in cooperative game theory, providing a fair and unique way to distribute the total value among the players in a coalition game. 

\begin{table*}[t]
\centering
\caption{Comparison of Key VFL contribution evaluation methods.}
\fontsize{8.5}{12}\selectfont
\begin{tabular}{c||>{\centering \arraybackslash}m{6.5cm}|c|c|c}
\Xhline{0.8pt}
Method                                                                       & {Paper}                                                                                                                                                      & Complexity & Approximation                                                             & Fairness                  \\ \hline
\multirow{5}{*}{\begin{tabular}[c]{@{}c@{}}Shapley Value\\ Based\end{tabular}} &Wang \cite{wang2019interpret}, Zhang et al. \cite{zhang2022data}                                                                                                     & $O(2^n)$   & /                                                                        &\Checkmark \\ \cline{2-5} 
                                                                               &PSO-EVFFS \cite{wang2019measure}                                                                                                                     & $O(n)$     & Monte Carlo Sampling            &\Checkmark \\ \cline{2-5} 
                                                                               &VerFedFL \cite{fan2022fair}                                                                                                                         & $O(n)$     & Monte Carlo  Sampling            &\Checkmark \\ \cline{2-5} 
                                                                               &FedValue \cite{han2021data}                                                                                                                         & $O(2^n)$   & /                                                                        &\Checkmark \\ \cline{2-5} 
                                                                               &DIG-FL \cite{wang2022efficient}                                                                                                                   & $O(n)$     & Training log and  Hassian matrix &\Xmark     \\ \hline
LOO Based                                                                      &HI-GAS \cite{sun2023hierarchical}, DIG-FL \cite{wang2022efficient}, Khan et al. \cite{khan2023incentive}, VFL-Market \cite{cui2024bargaining}                                                                            & $O(n)$     & /                                                                         &\Xmark     \\ \hline
Individual Based                                                               &VF-PS \cite{jiang2022vf}, MI-VFL \cite{xing2023distributed}, $\pi_{SS\_FS}, \pi_{H-FS}$ \cite{zhang2022secure}, FedSDG-FS \cite{li2023fedsdg}, FedSVM, FedRE \cite{ge2022failure}                                                          & $O(n)$     & /                                                                         &\Xmark     \\ \hline
\multirow{2}{*}{\begin{tabular}[c]{@{}c@{}}Interaction Based\end{tabular}}                                        & {MMVFL \cite{feng2020multi}, LESS-VFL \cite{castiglia2023less}, VFLFS \cite{feng2022vertical} , PSO-EVFFS \cite{zhang2022embedded}, Fed-EINI \cite{chen2021fed}} & $O(n)$   & /                                                                         &\Xmark     \\ \cline{2-5}
& Falcon \cite{wu2023falcon}, CRDP-FL \cite{zhao2022vertically}&$O(n^2)$ & / & \Xmark \\ \hline
\end{tabular}
\label{tab:complexity}
\end{table*}

However, following the definition of Shapley value in Equation \ref{eq:sv}, simply traversing all subsets of participants \cite{wang2019interpret,zhang2022data} to compute the Shapley value has exponential computation cost $\mathcal{O}(2^n)$ in retraining and testing. In VFL setting, this can result in a high communication overhead as parties need to exchange information about their models and datasets, leading to increased latency and resource usage. Therefore, a desired approach in Shapley value based contribution evaluation should consider to optimize on efficiency. 

Wang et al. \cite{wang2019measure} use situational importance (SI) as the utility function. The situational importance is the difference between what a feature contributes when it is at certain value and what it is expected
to contribute. For a set of features, the utility can be mathematically computed as, 
\begin{equation}
    \mathcal{U}(S)=\mathbb{E}[f|X_i=x_i,\forall i\in S]-\mathbb{E}[f],
\end{equation}
where $i$ is an index of features in $S$, $x_i$ is a specific feature value and $f$ is model prediction. An approximation algorithm with Monte-Carlo sampling is proposed to reduce the computational complexity to linear. Similarly, VerFedSV \cite{fan2022fair} extend the idea proposed in \cite{wang2019measure} to let clients’ contributions computed at multiple time points during the training process, using clients’ embeddings at different time-stamps, which also enhances efficiency. 

FedValue \cite{han2021data} addresses the problem of retraining models for different participant subsets by using model-independent utility functions, eliminating the need for retraining. It incorporates an information-theoretic metric Shapley-CMI to assess data values of multiple parties. More specifically, conditional mutual information (CMI) is metric is adopted as the utility function $\mathcal{U}$ in Equation \ref{eq:sv},
\begin{equation}
\mathcal{U}(i)=I(X_i;Y_t|X_t),
\end{equation}
where $X_i$ denotes the data of data party $i$, $Y_t$ and $X_t$ denotes the labels and data of the task party. The originally proposed method is of $O(2^n)$ complexity and it has been discussed that sampling and feature dimension reduction techniques are applicable to reduce computation cost.
%Additionally, the model-independent utility function could lead to fairness issues in asynchronous VFL.

On the other hand, Wang et al. \cite{wang2022efficient} approximate the Shapley value utilizing the training log the Hessian matrix of the loss and results in linear computation complexity.

% \begin{itemize}
%     \item (yz) Fair and efficient contribution valuation for vertical federated learning (SV + fair contribution + party level) \cite{fan2022fair}
%     \item (yz) Measure Contribution of Participants in Federated Learning (SV + feature-level, party level) deletion method \cite{wang2019measure}

%     \item (cj) Data valuation for vertical federated learning: An information-theoretic approach (SV + privacy-aware + feature-level) \cite{han2021data}
%     \item (cj) Interpret federated learning with shapley values (SV + interpretable VFL + feature-level) \cite{wang2019interpret}

%     \item (cy) Efficient participant contribution evaluation for horizontal and vertical federated learning (SV + LOO + efficient compute + party-level) \cite{wang2022efficient}

%     \item (cj) Data Pricing in Vertical Federated Learning (SV + feature-level) \cite{zhang2022data}
    
% \end{itemize}

\subsubsection{Leave-one-out Based}

The Leave-one-out (LOO) method typically involves excluding the contributions of each party/feature, one at a time, and analyzing the impact this has on the global model performance. The principle behind this method is to elucidate the utility of each player's contribution towards the utility function of the model. If the performance increases when certain data is left out, it indicates that this data may be misleading or detrimental to the overall learning process.

Mathematically, this can be expressed as:
\begin{equation}
v(i)=\mathcal{U}(S\cup \{i\})-\mathcal{U}(S),
\label{eq:loo}
\end{equation}
where $v(i)$ here denotes the contribution of participant $i$.

The computation complexity of LOO based approach is $\mathcal{O}(n)$. As the calculation of LOO is straightforward and efficient, it is valuable for scenarios where numerous parties and features are involved and retraining is impractical \cite{sun2023hierarchical,wang2022efficient,cui2024bargaining}. In \cite{cui2024bargaining}, the data party's contribution is quantified by the performance gain that its features bring to the task party's model, and this quantification is used as the basis for the bargaining process to determine the payment that the task party will offer for the use of these features. 

However, it lacks fairness properties in Shapley value and thus the measurement may not be proper in some conditions. LOO fails to accurately evaluate the contribution of participants in the presence of duplicates or highly similar ones. The LOO-based metric can be sensitive to the order in which players are considered. Different orderings may lead to different contributions for the same player. Moreover, LOO may not be additive, meaning that the sum of individual contributions may not equal the value of the grand coalition. However, as fairness is subjective, it can be defined in other forms, which may be satisfied by LOO. For example, Khan et al. \cite{khan2023incentive} adapt LOO using Talmud’s division rule, which ensures the fairness defined based on the Bankruptcy Problem.

% \begin{itemize}
%     \item (cj) Incentive Allocation in Vertical Federated Learning Based on Bankruptcy Problem (bankruptcy problem + feature-level) \cite{khan2023incentive}

%     \item (cy) Hierarchical Federated Learning Incentivization for Gas Usage Estimation (LOO + contribution evaluation + payment allocation + party level) [new category]\cite{sun2023hierarchical}

%     \item (cj) Efficient participant contribution evaluation for horizontal and vertical federated learning (SV + LOO + efficient compute + party-level) \cite{wang2022efficient}
% \end{itemize}

\subsubsection{Individual Based}
%\subsubsection{Mutual Information Based}

Instead of evaluating contribution through the relationship between a participant and a coalition, individual methods seek to measure the contribution of each party/party based on the diversity and usefulness of each participant's individual data. In this group, the measurements of contribution mainly include,  mutual information-based \cite{jiang2022vf,xing2023distributed}, Gini score based \cite{zhang2022secure,li2023fedsdg,ge2022failure}, and counterfactual distance based \cite{chen2022evfl}. A higher entropy or larger distance corresponds to more randomness and typically, a bigger contribution to the model update. 

VF-PS \cite{jiang2022vf} aims at selecting the subset of features that jointly preserve the greatest amount of mutual information with the label. However, such a framework requires computation over the data of participants. To achieve so in a privacy-preserving way, Jiang et al. \cite{jiang2022vf} incorporate homomorphic encryption. Xing et al. \cite{xing2023distributed} argue that when measuring the mutual information between features and labels, the subset of features that maximize the term may be those that are not important individually, but combinatorially. While focusing on the combination can lead to the missing of individual important features. Xing et al. \cite{xing2023distributed} adapt an adversarial game to deal with this problem. It maximizes
the mutual information between the selected feature subset and the labels, while also minimizing the mutual information between remaining feature subsets and the labels. 

Gini impurity is a concept used in the context of decision trees and machine learning classification algorithms. In decision tree algorithms, the Gini impurity is a measure of how often a randomly chosen element would be incorrectly classified. It is used to evaluate the quality of a split in the data during the construction of a decision tree. A node with low Gini impurity is considered "pure," meaning that most of the elements in that node belong to the same class. Therefore it can be used as the measurement of the feature importance score. Zhang et al. and Ge et al. \cite{zhang2022secure,ge2022failure} calculate the Gini impurity of each individual feature w.r.t. the labels to measure the importance of a feature. Li et al. \cite{li2023fedsdg} indicate that directly using Gini impurity as feature importance is inappropriate as it cannot take into account the specific VFL models. Therefore, they propose FedSDG-FS, which takes Gini impurity as the initialization of features' importance and learns the final indicators of whether a feature should be selected by jointly training an interactive layer with the VFL model that performs the downstream task.

Computing entropy-related measurements, particularly in a distributed and privacy-preserving manner, may require complex methods that come with their computational and implementation costs. Therefore, entropy-based measurements has the drawback that extra privacy-preserving efforts \cite{jiang2022vf,li2023fedsdg,zhang2022secure} may need to be made to ensure data privacy. 

EVFL \cite{chen2022evfl} adapts the idea of counterfactual explanation to the setting of explainable VFL and proposes a counterfactual instance-based party importance metric. The key idea of counterfactual explanation is to provide insights into why a particular decision was made by the model through highlighting the factors or conditions that, if changed, would lead to a different outcome. Counterfactual instances $c^m$ are generated separately on each client $m$, and the distance between the counterfactual instance and query instance $x^m$ is used to measure the importance of party $m$. Mathematically, the importance score can be formulated as
\begin{equation}
    |(x^m-c^m)/x^m|.
\end{equation}

% \begin{itemize}
%     \item (cj) VF-PS: How to Select Important Participants in Vertical Federated Learning, Efficiently and Securely? (privacy-aware + vertically federated participant selection + feature-level + MI) \cite{jiang2022vf}

%     \item (cj) Distributed Model Interpretation for Vertical Federated Learning with Feature Discrepancy (MI + feature-level + privacy-aware) \cite{xing2023distributed}
% %\end{itemize}

% %\subsubsection{Gini-impurity Based}
% %\begin{itemize}
%     \item (yz) Secure Feature Selection for Vertical Federated Learning in eHealth Systems (privacy-aware + Gini-based + feature selection) \cite{zhang2022secure}

%     \item (cy) FedSDG-FS: Efficient and Secure Feature Selection for Vertical Federated Learning (gini + privacy-aware + feature-level + feature selection) \cite{li2023fedsdg}

%     \item (cj) Failure Prediction in Production Line Based on Federated Learning: An Empirical Study (gini + feature selection) \cite{ge2022failure}

% \end{itemize}

\subsubsection{Interaction Based}
Another line of methods of contribution evaluation in VFL requires the interaction of certain individual's data, using measurements such as learnable feature weights \cite{feng2020multi,castiglia2023less,feng2022vertical}, correlation coefficient \cite{zhao2022vertically,wu2023falcon}, and node split time \cite{zhang2022embedded,chen2021fed}.

As feature/party contribution evaluation can be interpreted as identifying the influence of features/parties, it has great similarity with sparse learning. Sparse selection is a process used in machine learning and statistics to identify and retain only the most relevant features or variables from a larger set of features. The goal is to reduce the dimensionality of the data by selecting a subset of informative features while discarding irrelevant or redundant ones. Feng et al. \cite{feng2020multi} follow the idea of sparse learning and uses a transformation matrix to project data to new space. The L2 norm of corresponding row of the matrix is considered as the importance score of the feature. To ensure sparsity, it includes a regularization as penalty term. Instead of conducting sparsifying on original features, in VFLFS \cite{feng2022vertical}, feature data is mapped into a lower-dimensional space in a way that preserves certain relationships or properties to measure its importance. VFLFS utilizes auto-encoder to project features into embedding space and measure the importance of features via a transformation matrix similar to \cite{feng2020multi}. Group lasso has also been widely used as a technique for sparsifying. It extends the lasso regularization to scenarios where features can be naturally grouped together. It introduces an additional penalty term that encourages sparsity not only at the individual feature level but also at the group level. Castiglia et al. \cite{castiglia2023less} propose LESS-VFL that adapts the idea of group lasso to measure the importance of features. However, directly applying group lasso requires the feature to be selected by the server from a global view, and thus the parties and server need to exchange data embeddings and partial derivatives every iteration of training, which is communication inefficient. LESS-VFL deals with this by performing server-side and party-side group lasso hierarchically, where the goal of party-side group lasso is to select significant features that match the significant components selected by the server. 

Correlation coefficient has also been used to measure the importance of features \cite{zhao2022vertically, wu2023falcon}. For example, Zhao et al. \cite{zhao2022vertically} use stability feature selection to traverse the distributed data feature set across parties and filter out the less correlated features to the prediction. It then adopts Pearson’s correlation coefficient to merge feature sets on feature-partitioned distributed data. To protect privacy, DP noise is injected into the data.

%\textbf{Counterfactual Explanation Based.} 

Tree-based VFL models has its uniqueness in measuring the contribution of features, which can be done by counting the node split time \cite{zhang2022embedded,chen2021fed}. PSO-EVFFS \cite{zhang2022embedded} is proposed under the protocol of SecureBoost \cite{cheng2021secureboost}, in which participants jointly train multiple XGBoost tree models and count the average number of split times for each feature. The feature importance is measured in a two-stage manner. First, on the side of each data party, it marks features as irrelevant if the split times are zero. Irrelevant features are removed in subsequent the stage. In the second stage, the multi‐participant cooperative particle swarm optimization (PSO) algorithm is used to search the optimal feature subset by taking the remaining features as full set.

\subsection{Tasks Involving Contribution Evaluation}\label{Task}

In this section, we divide VFL contribution evaluation into 5 subtasks. Various tasks are performed at different phases of the lifecycle, each contributing to the development and operation of a VFL model. Table \ref{tab:relation} delineates the occurrence of tasks within the VFL process, offering a structured understanding of its lifecycle. Each task's significance is emphasized in the context of the entire lifecycle, with empty entries indicating the potential for future research where current literature is lacking. The data preprocessing phase encompasses tasks such as feature selection and payment allocation, during which participants determine the gradients of collaboration. The VFL training phase includes tasks related to feature selection, incentive mechanisms, and payment allocation, wherein the involved parties and their contributions are carefully monitored and adjusted in accordance with the optimization process of the VFL model. As for the VFL inference phase, it may involve tasks related to contribution evaluation, interpretable VFL, and payment allocation, where the efficacy of the model is evaluated.

\begin{table*}[t]
\centering
\caption{Overview of task presence across different phases of the VFL lifecycle.}
\fontsize{8.5}{12}\selectfont
\label{tab:relation}
\begin{tabular}{>{\centering\arraybackslash}m{3cm}||>{\centering\arraybackslash}m{4cm}|>{\centering\arraybackslash}m{5cm}|>{\centering\arraybackslash}m{4cm}}
\Xhline{0.8pt} 
                           % & \multicolumn{3}{c}{\textbf{VFL phase}} \\
   \diagbox{Task}{Lifecycle}                        & \multicolumn{1}{c|}{Data Preprocessing} & \multicolumn{1}{c|}{VFL Training} & \multicolumn{1}{c}{VFL Inference} \\ \hline
Feature Selection & VF-PS \cite{jiang2022vf}, FedSDG-FS \cite{li2023fedsdg} & FedSDG-FS \cite{li2023fedsdg}, VerFedSV \cite{fan2022fair}, LESS-VFL \cite{castiglia2023less}, $\pi_{SS\_FS}, \pi_{H-FS}$ \cite{zhang2022secure}, FedSVM, FedRF \cite{ge2022failure}, VFLFS \cite{feng2022vertical}, PSO‐EVFFS \cite{zhang2022embedded}, MMVFL \cite{feng2020multi}, CRDP-FL \cite{zhao2022vertically} & \Xmark \\ \hline
Incentive Mechanism & \Xmark & Khan et al. \cite{khan2023incentive} & \Xmark  \\ \hline
Interpretable VFL & \Xmark  & \Xmark  & Wang \cite{wang2019interpret}, EVFL \cite{chen2022evfl}, MI-VFL \cite{xing2023distributed}, Fed-EINI \cite{chen2021fed}, Falcon \cite{wu2023falcon}      \\ \hline
Payment Allocation & HI-GAS \cite{sun2023hierarchical} & HI-GAS \cite{sun2023hierarchical}, VFL-Market \cite{cui2024bargaining}  & \Xmark      \\ \hline

\end{tabular}
\end{table*}

\subsubsection{Feature Selection}
In the realm of VFL, the task of feature selection plays a pivotal role in enhancing the collaborative model-building process across decentralized data sources. The primary goal of feature selection in VFL is to identify and prioritize relevant features from each participant's local dataset, contributing to the creation of a more concise and effective global model. Contribution evaluation, in this context, is intimately tied to assessing the significance of features contributed by each participant. The challenge lies in devising a contribution evaluation mechanism that not only accurately measures the impact of features on the global model but also efficiently selects these significant features as a prior of or at least simultaneously to the training process of the global model.

One line of works addresses feature selection within the framework comprising a central server/the task party and multiple other client participants \cite{castiglia2023less,jiang2022vf,li2023fedsdg,zhang2022secure,zhang2022embedded}. The selection process unfolds with the server controlling the objective of selection. The challenges lie in how to perform privacy preserving optimization of the selection process. In \cite{castiglia2023less}, the selection contains three stages. Firstly, models undergo pre-training in a standard VFL environment. Secondly, the server identifies the crucial embedding components among the clients' sets. Thirdly, each party aims to align its model's significant embedding components while discarding non-significant features. 

While in \cite{feng2022vertical,zhao2022vertically,zhang2022secure}, they deal with the scenario where feature selection is performed on each FL participant. Feng \cite{feng2022vertical} proposes VFLFS, which operates by jointly training a classifier model across all participating parties. Additionally, it imposes row sparsity on the weight matrix connecting the input layer to the second layer of the local model at each party to represent feature importance.

% \cite{fan2022fair,jiang2022vf,zhang2022secure,li2023fedsdg,ge2022failure,feng2022vertical,zhang2022embedded,li2023towards,feng2020multi,zhao2022vertically,huangEfficiencyboostingClientSelection2020a,cheungVerticalFederatedPrincipal2022a,liFeatureSelectionData2018,tibshirani1996regression,lemhadri2021lassonet,dinh2020consistent,gorishniy2021revisiting}

\subsubsection{Interpretable VFL}
Interpretable VFL represents another important task where the emphasis is on enhancing the transparency and comprehensibility of the collaborative model. The goal here is to provide interpretable insights into the contributions of individual participants/features. Interpretable FL is closely linked to contribution evaluation due to its capacity to elucidate prediction outcomes, aid in model debugging, and offer insights into the specific contributions made by individual data owners or data samples. This understanding is pivotal for the equitable allocation of rewards, which serves to incentivize proactive and dependable participation in FL endeavors. 

The very first challenge for interpretable VFL is privacy issue. Traditional model interpretation methods cannot be directly applied in VFL due to privacy concerns. Wang  \cite{wang2019interpret} adapts SHAP \cite{lundberg2017unified} for feature interpretation in VFL. It deals with privacy issue by passing feature IDs across parties to turn on and turn off specific features for calculating Shapley value, which is used as the measure of feature importance. Chen et al. \cite{chen2021fed} address the task of improving the interpretability of vertical decision tree ensembles in FL. The interpretability issue is tackled by proposing a framework named Fed-EINI, which allows secure and interpretable inference. This involves disclosing the meanings of features from the Host party to the Guest party while ensuring the privacy of the decision paths by using encryption techniques. The importance of features is measured by the weight of corresponding nodes. Falcon \cite{wu2023falcon} introduces a framework, Falcon-INP, that supports interpretable methods like LIME \cite{ribeiro2016should} in a decentralized setting, allowing interpretability calculations that ensure each party's data remains private. The Falcon-INP framework ensures that the intermediate information computed on the parties' data is in secure formats. Xing et al. \cite{xing2023distributed} deal with the local-global discrepancy and local-local discrepancy caused by feature misalignment in interpretable VFL. It adjusts the local importance of features to align with their global importance using the law of total probability, ensuring that the selected features represent their true contribution to the model's performance. Chen et al. \cite{chen2022evfl} propose a credible federated counterfactual explanation (CFCE) method to perform feature interpretation in VFL.

%\cite{wang2019interpret,xing2023distributed,chen2021fed,wu2023falcon}
%\subsubsection{Plain Contribution Measuring}
%The task of plain contribution measuring in VFL revolves around the efficient calculation of the Shapley value, a cooperative game theory concept used to fairly allocate the total contribution of each participant in the collaborative learning process. The goal is to accurately quantify the individual impact of each participant on the model's performance, promoting fairness in contribution attribution. Contribution evaluation, in this case, is synonymous with Shapley value calculation and demands an efficient algorithmic approach to handle the complexity of federated settings \cite{wang2019measure,han2021data,wang2022efficient,zhang2022data,sun2023hierarchical}. The required properties of contribution evaluation in this scenario include efficiency and accuracy in Shapley value estimation, ensuring a precise reflection of each participant's unique contribution in VFL.

\subsubsection{Incentive Mechanism Design}
Incentive mechanism represents a distinctive task in VFL, focusing on motivating active participation and collaboration among decentralized entities. The goal is to design mechanisms that incentivize participants to contribute valuable information without compromising privacy or security. Contribution evaluation, within the context of incentive mechanisms, becomes a critical component for determining the rewards or incentives allocated to each participant based on their contributions. The properties essential for contribution evaluation in this scenario encompass not only fairness and efficiency but also strategic considerations, ensuring that the incentive mechanism aligns with the overarching goal of fostering collaboration while safeguarding individual privacy and interests in the VFL ecosystem \cite{khan2023incentive}.
\subsubsection{Payment Allocation}
Payment allocation is an emerging task in VFL that centers around the equitable distribution of payments among participating entities. The goal is to establish a fair and transparent payment system that accurately reflects the contributions of each participant to the collaborative learning process. Contribution evaluation, in the context of payment allocation, is synonymous with the assessment of each participant's impact on the global model's performance, forming the basis for determining their respective share of the overall payment. The required properties for contribution evaluation in payment allocation extend to encompass fairness, efficiency, and transparency, ensuring an equitable distribution of rewards that aligns with the collaborative nature of VFL while upholding principles of fairness and accountability \cite{sun2023hierarchical}. With the development of data/model markets \cite{cui2024auction,cui2024bargaining,zheng2022fl,cong2022data,pei2020survey}, the pricing of features is also studied in the context of VFL. Cui et al. \cite{cui2024bargaining} adapt bargaining mechanism under one-on-one VFL setting to perform feature trading. It proposes performance gain-based pricing and introduces bargaining models that can fairly reward the features of the task party and reach equilibrium that optimizes the parties’ objectives.

% \subsubsection{Not Specified}
% \cite{catonFairnessMachineLearning2020,blackLeaveoneoutUnfairness2021}

%\input{subfiles/0_taxonomy_tree}
\section{Open challenges and Future directions}
\label{sec:cha}
% (May refer to the following methods in general setting when describing method)
% \begin{itemize}
%     \item (cy) A Performance-Driven Benchmark for Feature Selection in Tabular Deep Learning (feature-level + lasso based + feature selection) \cite{cherepanova2023performance}: no open challenge in discussion.
    
%     \item (cy) Regression Shrinkage and Selection via the Lasso (feature level + lasso based + feature selection) \cite{tibshirani1996regression}
%     \item (cy) LassoNet: A Neural Network with Feature Sparsity (feature level + lasso based + feature selection) \cite{lemhadri2021lassonet}
%     \item (cy) Consistent feature selection for analytic deep neural networks (feature level + lasso based + feature selection) \cite{dinh2020consistent}
%     \item (cy) Revisiting deep learning models for tabular data (feature level + attention based + feature selection) \cite{gorishniy2021revisiting}
%     \item (cy) least core based

%     \item (yz) Feature Selection: A Data Perspective\cite{liFeatureSelectionData2018}
%     \item (yz) SHAFF: Fast and consistent SHApley eFfect estimates via random Forests (SV + tree)\cite{benardSHAFFFastConsistent2021}
%     \item (yz) Fairness in Machine Learning: A Survey (ML)\cite{catonFairnessMachineLearning2020}

%     \item (yz) Leave-one-out Unfairness (ML)\cite{blackLeaveoneoutUnfairness2021}
    
% \end{itemize}
 % In this section, we discuss some of the major open challenges faced in VFL contribution evaluation from the perspectives of fairness of contribution evaluation method, evaluation method design and others.
The evaluation of contributions throughout the lifecycle of VFL is a multifaceted challenge. The interplay between privacy, communication overhead, and fairness forms a triad of challenges that must be balanced to form a healthy ecosystem of VFL, which includes data generation, usage, and valuation.

\subsection{Privacy and Security} 
The training diagram of VFL is designed to protect user privacy by not collecting private data. However, there are still privacy concerns that need to be addressed in the lifecycle of VFL training, and there is the possibility that the contribution evaluation be the victim of attackers. For example in the model training phase, threats including model attacks from semi-honest/malicious participants are worth noticing before evaluating the contribution. According to different attacking targets, model performance may be undermined in the scenario where malicious parties exist, leading to the false evaluation. And semi-honest parties could post threats on privacy during the evaluation process. Therefore, the design of VFL contribution evaluation should not only consider processing with the secured data but also the method itself should be secure. Thus the defence mechanisms should be clarified along with the contribution evaluation methods. Beyond conventional safeguards such as traditional differential privacy, homomorphic encryption and secure multi-party computing, advanced techniques like zero-knowledge proof (ZKP) \cite{abs-2305-04507, ZhangFZS20, RuckelSH22} and trusted execution environment (TEE) \cite{ZhangLLCLGC23, HuangLCZSCY21, MoHKMPK21} can also be used to ensure the security of the evaluation process from the algorithm and hardware levels respectively.

\subsection{Communication Overhead} Communication overhead represents a critical consideration in the development of contribution evaluation methodologies. Certain approaches necessitate the frequent transfer of substantial data volumes, including embeddings or derived matrices, between parties and the agent responsible for calculating contributions. This can introduce significant bottlenecks, impeding the efficiency of the process in terms of both time and resource allocation. As the number of parties or the volume of data grows, contribution evaluation methods with high communication overhead struggle to manage the increased system load, with proportional increases in cost or decreases in performance. Furthermore, communication overhead can also have implications for privacy and security. More frequent or voluminous data exchanges increase the risk of interception or unauthorized access. Consequently, addressing communication overhead is imperative to realize the practical implementation of a feasible and sustainable VFL evaluation system. The one-shot VFL method designed by Sun et al. \cite{00020YNLZX0R23} allow participants to perform two upload and one download operations during the training phase, markedly decreasing communication costs and frequency. Zhang et al. \cite{ZhangGDGBPH21} propose an asynchronous stochastic quasi-Newton framework for VFL, which can enhance convergence speed over SGD-based by approximating Hessian information, thus significantly diminishing the requisite number of communication rounds. Fu et al. \cite{abs-2207-14628} have crafted an efficient VFL training framework that employs cache-enabled local update technology to improve system efficiency and introduces the abstraction of a workset table to slash communication costs. Despite these advancements, optimization of vertical contribution evaluation strategies, with a specific focus on communication overhead reduction, remains largely unaddressed, signifying a gap in current research efforts. While an existing contribution evaluation method \cite{castiglia2023less} does acknowledge the importance of optimizing communication overhead, this aspect has yet to be prioritized adequately. Consequently, addressing communication overhead within the contribution evaluation process emerges as a critical and burgeoning objective for optimization. This gap underscores the necessity for dedicated research efforts aimed at developing methodologies that not only assess contributions effectively but also do so with an optimized communication footprint, thereby enhancing the overall efficiency and sustainability of VFL systems.

\subsection{Fairness}

\subsubsection{The comparison of contribution evaluation methods}
We notice that there is no consensus nor benchmarks on how to compare the fairness of different contribution evaluation methods even for the same phase in the lifecycle of VFL, making the comparison of contribution evaluation methods difficult. Therefore, it's necessary to devise evaluation metrics \cite{chai2024survey}, hyper-parameters fine-tuning strategies, and other essential experimental settings to evaluate the fairness of contribution evaluation methods.

%\subsubsection{Incentive mechanism design} 
%A healthy and self-sustainable VFL ecosystem should motivate diverse entities to collaborate and contribute their private data or computational resources to a common learning task. Current works on VFL contribution evaluation mostly study the problem from a global perspective and lack the exploration of incentive mechanism design. Fairness is a crucial aspect of the incentive mechanism. Parties may contribute less valuable data or resources if they feel that the incentive system does not reward them adequately. It is important to design mechanisms that discourage minimal or subpar contributions. In the meanwhile, there is a risk that parties may find ways to manipulate the contribution evaluation to receive undue incentives. Robust detection and penalization of such behavior are essential to maintain the integrity of the VFL system. Another challenge is that contributions may change over time as parties update their data or as the model's requirements evolve. Incentive mechanisms need to be flexible to accommodate these changes and continue rewarding parties fairly.

\subsubsection{A general contribution evaluation framework for the whole lifecycle of VFL}
As is mentioned in the previous sections, contribution evaluation takes part in each part of the lifecycle in the VFL training process, i.e. in the data preprocessing phase, contribution evaluation mainly focuses on feature selection and dimensionality reduction \cite{jiang2022vf,han2021data,li2023fedsdg}; for model training phases, contribution evaluation serves the purposes of optimizing model performance and measuring the real contribution of all parties \cite{fan2022fair,cui2024bargaining,castiglia2023less} and interpretability of model prediction at model inference phase \cite{xing2023distributed,chen2021fed,wang2019interpret}. However, the evaluation methods are independent and separate according to different phases, thus lack of the ability to depict the whole picture of the VFL diagram nor did the contribution is guaranteed to be fairly measured in subsequent phases. It is challenging but necessary to devise a general framework/metric that measures the overall contribution of each party in a VFL lifecycle.

\subsection{Multi-objective Optimization}
The necessity to simultaneously optimize privacy, communication overhead, and fairness presents a significant multi-objective optimization challenge \cite{gu2023theoretical, kang2023optimizing,zhang2023trading,zhang2022no}. Each of these objectives is essential for the success and widespread adoption of VFL, but they often have conflicting requirements that can make their simultaneous optimization a complex task. 

Privacy-enhancing techniques, such as encryption and secure multi-party computation \cite{goldreich1998secure}, come at the cost of increased computational complexity and communication overhead . As the privacy guarantees become stronger, the amount of data and the number of messages that need to be exchanged often increase, leading to higher communication costs. Optimizing one without affecting the other negatively is a critical concern in VFL. Fair contribution evaluation requires a certain level of transparency into the data or model parameters of each participant \cite{fan2022fair}, which can conflict with the goal of privacy preservation. As privacy measures become more stringent, the ability to accurately assess and attribute contributions can become obscured, potentially leading to fairness issues. Reducing communication overhead often means limiting the amount of information exchanged between parties \cite{konevcny2016federated}. However, this could impact the granularity at which contributions can be measured and assessed, potentially compromising the fairness of the evaluation process. Ensuring that each party's contribution is fairly evaluated may necessitate more frequent or detailed communication, which can exacerbate the overhead.

\subsection{Data Valuation} 
Data pricing, asset management, trading, and market dynamics \cite{pei2020survey,cui2024auction,cui2024bargaining,zheng2022fl} are promising areas for future research and practical application and is closely related to contribution evaluation in VFL setting. This integration can lead to the development of a comprehensive ecosystem where data, computational resources, and models are valued, exchanged, and utilized in a fair and efficient manner.

The concept of data as an asset can be extended to VFL, where data and models become tradable commodities. This requires the development of asset management strategies that align with the principles of VFL, such as data privacy and model security. By treating data and models as assets, VFL can establish a framework for valuing and managing these resources, which can then be used to create investment and trading opportunities within the VFL ecosystem. This can lead to the emergence of new business models and market dynamics centered around data and machine learning assets.

Data pricing is a critical component in the data economy, and its application in VFL can help quantify the value of data contributed by participants. VFL systems can implement pricing mechanisms that incentivize data sharing while ensuring that the contributors are fairly compensated. This can lead to a more vibrant data market, where the quality, relevance, and utility of the data are factored into the pricing model, thus encouraging the sharing of high-quality data that benefits the entire VFL ecosystem. Designing incentive mechanisms that reward participants for sharing high-quality data while maintaining privacy could be a key area of research. It is also promising to explore the feasibility and design of a marketplace for VFL resources, including the development of trading protocols, auction methods, bargaining methods \cite{cui2024bargaining}, and contract theories tailored for VFL. Moreover, methods for trading data and models that preserve privacy while allowing for transparent and fair evaluation of contributions should also be developed.

% \subsection{Others}
% In this part we will discuss other challenges for contribution evaluation. Besides from the previous aspects, challenges including data heterogeneity and non-linearity are also commonly encountered in VFL scenario, which may affects the evaluation of the contribution among different parties. Therefore the evaluation methods should also take this into consideration

\section{Conclusion}
\label{sec:conc}
In conclusion, this survey paper has provided a comprehensive analysis of the critical role of contribution evaluation in VFL, highlighting its importance in ensuring privacy-preserving, fair, and effective collaboration among data owners. We have presented a detailed taxonomy that categorizes contribution evaluation methods based on their objectives, subjects, techniques, and tasks, offering a structured framework for understanding and assessing contributions throughout the VFL lifecycle. We also present open challenges in this area from the perspectives of privacy and security, communication overhead, fairness, multi-objective optimization, and data valuation. We hope the survey will inspire further research and development of more equitable and adaptive contribution evaluation mechanisms.

\section*{Acknowledgments}The research work described in this paper was supported by Hong Kong Productivity Council (Grant No.HKPC22EG01-B). It was partially conducted in JC STEM Lab of Data Science Foundations funded by The Hong Kong Jockey Club Charities Trust. This work was also supported by NSFC Grants no. 72071125, 72031001, and 62376118.

\bibliographystyle{IEEEtran}
\bibliography{refs}

\end{document}